\documentclass[times, review, 10pt]{elsarticle} 

\makeatletter
\def\ps@pprintTitle{%
 \let\@oddhead\@empty
 \let\@evenhead\@empty
 \let\@oddfoot\@empty
 \let\@evenfoot\@empty
}
\makeatother

\usepackage{amssymb}
\usepackage{amsmath}

\usepackage{url}  
\usepackage{multirow}
\usepackage{booktabs}  
\usepackage{subcaption}
\usepackage{amsmath}
\usepackage{xcolor}
\usepackage{algorithm}      
\usepackage{algpseudocode}  
\usepackage{soul} 

\begin{document}

\begin{frontmatter}



\title{The Impact of the Single-Label Assumption\\in Image Recognition Benchmarking} 

\author[label1,label2]{Esla Timothy Anzaku\corref{cor1}}
\ead{eslatimothy.anzaku@ugent.be}
\cortext[cor1]{Corresponding author.}

\author[label1,label2]{Seyed Amir Mousavi}

\author[label3]{Arnout Van Messem}

\author[label1,label2]{Wesley De Neve}

\affiliation[label1]{organization={Department of Electronics and Information Systems, Ghent University},
            addressline={Technologiepark-Zwijnaarde 126},
            city={Ghent},
            postcode={9052},
            state={},
            country={Belgium}}

\affiliation[label2]{organization={Center for Biosystems and Biotech Data Science, Ghent University Global Campus},
            addressline={119 Munhwa-ro 119-5, Yeonsu-gu},
            city={Incheon},
            postcode={21985},
            state={},
            country={South Korea}}

\affiliation[label3]{organization={Department of Mathematics, University of Liège},
            city={Liège},
            postcode={4000}, 
            country={Belgium}}

\begin{abstract}
Deep neural networks (DNNs) are typically evaluated under the simplifying assumption that each image corresponds to a single correct label. However, many images in widely used benchmarks such as ImageNet contain multiple valid labels, resulting in a mismatch between the evaluation assumptions and the actual complexity of visual data. Consequently, DNNs are penalized for predicting valid labels that differ from the assigned single-label ground truth. This mismatch between model predictions and the ground-truth label may substantially contribute to reported performance gaps—most notably, the widely discussed $11$-$14$\% drop in top-1 accuracy on ImageNetV2, a test set designed to replicate ImageNet.
 This observation raises critical questions about whether the reported drop reflects genuine limitations in model generalization or is instead an artifact stemming from restrictive evaluation metrics.
In this study, we rigorously quantify the impact of multi-label characteristics on observed accuracy discrepancies. To evaluate the multi-label prediction capability (MLPC) of single-label models, we introduce a variable top-$k$ evaluation strategy, where $k$ equals the number of valid labels present in each image. This approach is based on the premise that models trained with single-label supervision will nonetheless rank multiple correct labels near the top when present. Our analysis, based on 315 ImageNet-trained models, confirms that traditional top-1 metrics disproportionately penalize valid but secondary labels. To better aggregate model behavior under multi-label evaluation, we further propose Aggregate Subgroup Model Accuracy (ASMA) as a metric. Importantly, we demonstrate substantial variability among models in their ability to accurately rank multiple correct labels, with certain models consistently exhibiting superior MLPC. Under the proposed evaluation method, the apparent accuracy gap between ImageNet and ImageNetV2 is substantially reduced, indicating that conventional single-label accuracy metrics can exaggerate differences and obscure true model capabilities.
To further isolate multi-label recognition performance from potential contextual biases, we introduce PatchML, a synthetic dataset composed of systematically structured combinations of object patches. Evaluations on PatchML reveal that DNNs trained solely with single-label annotations on ImageNet inherently possess a noteworthy capability to recognize multiple objects simultaneously.
Collectively, our findings highlight critical shortcomings in conventional single-label evaluation frameworks, suggesting that they underestimate the actual recognition capabilities of contemporary DNNs and risk misleading conclusions about model generalization. Given that real-world images routinely contain multiple objects, aligning evaluation methods with this inherent complexity is essential. Our work emphasizes the necessity of adopting multi-label-aware evaluation protocols to better reflect real-world visual recognition tasks, ultimately contributing to the development of more reliable and trustworthy neural network models.
\end{abstract}


\end{frontmatter}

\section{Introduction}

\begin{figure*}[htbp]
    \centering
    \begin{minipage}[b]{0.3\textwidth}
        \includegraphics[width=\textwidth]{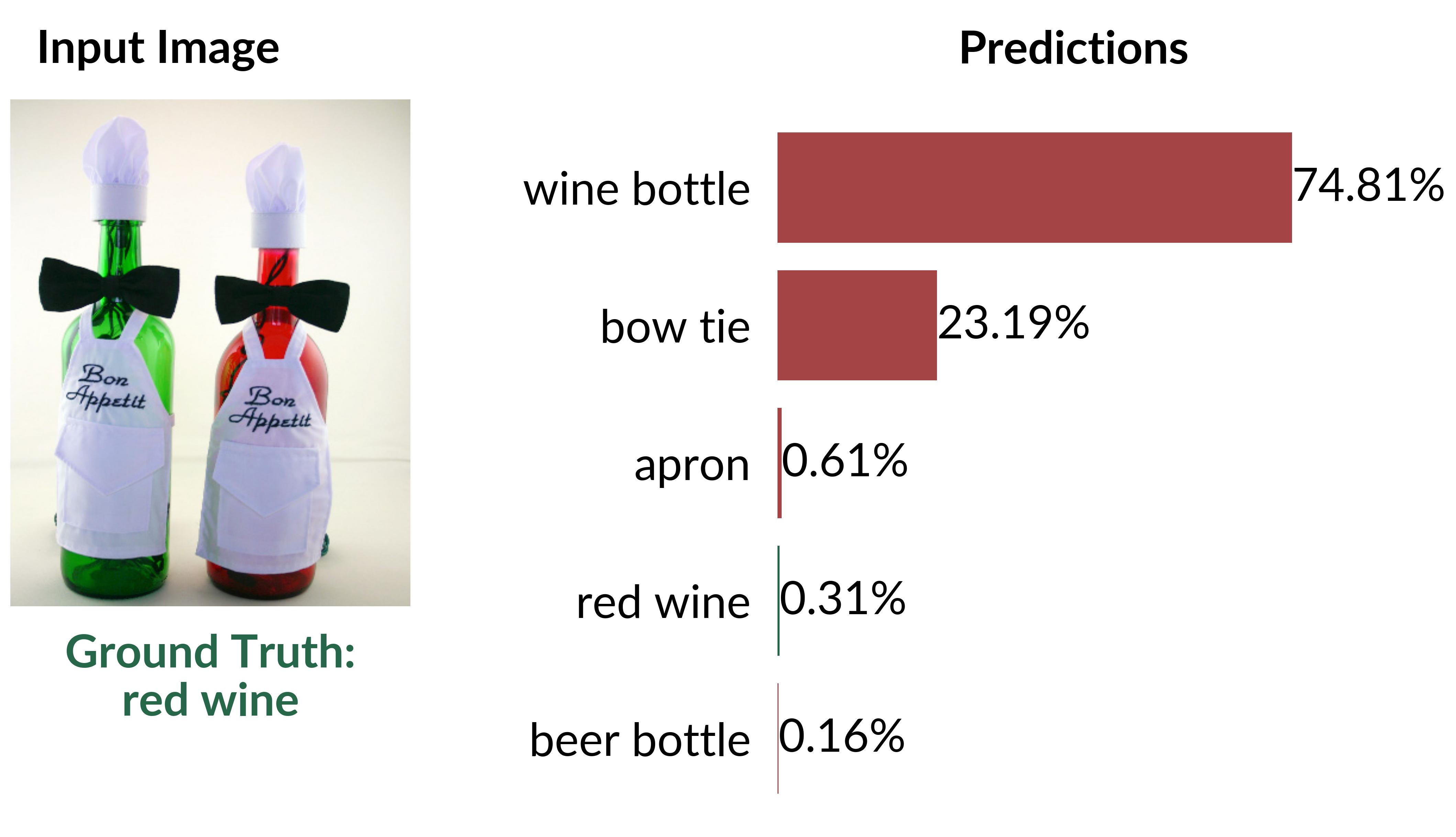}
    \end{minipage}
    \hfill 
    \begin{minipage}[b]{0.3\textwidth}
        \includegraphics[width=\textwidth]{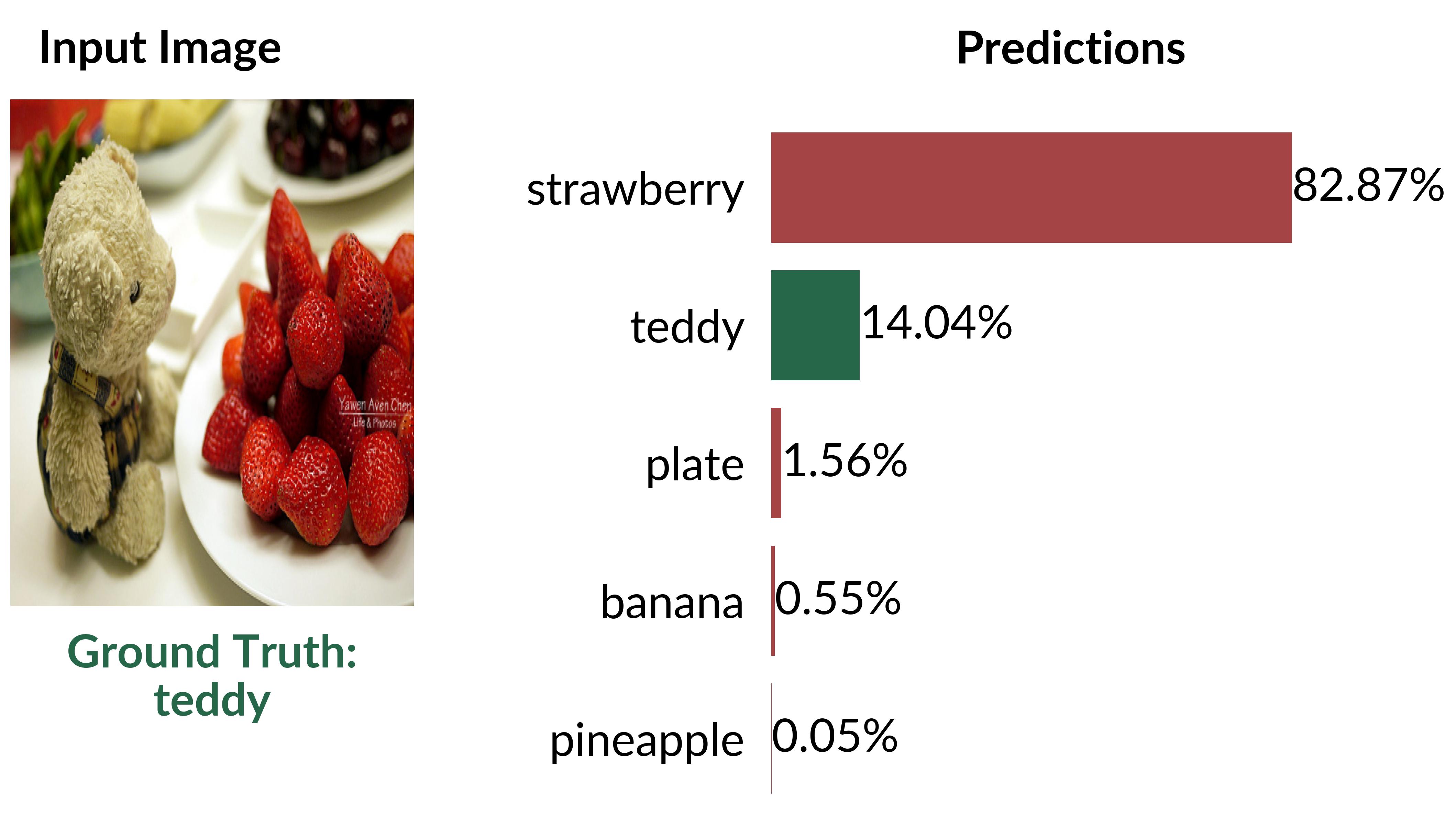}
    \end{minipage}
    \hfill 
    \begin{minipage}[b]{0.3\textwidth}
        \includegraphics[width=\textwidth]{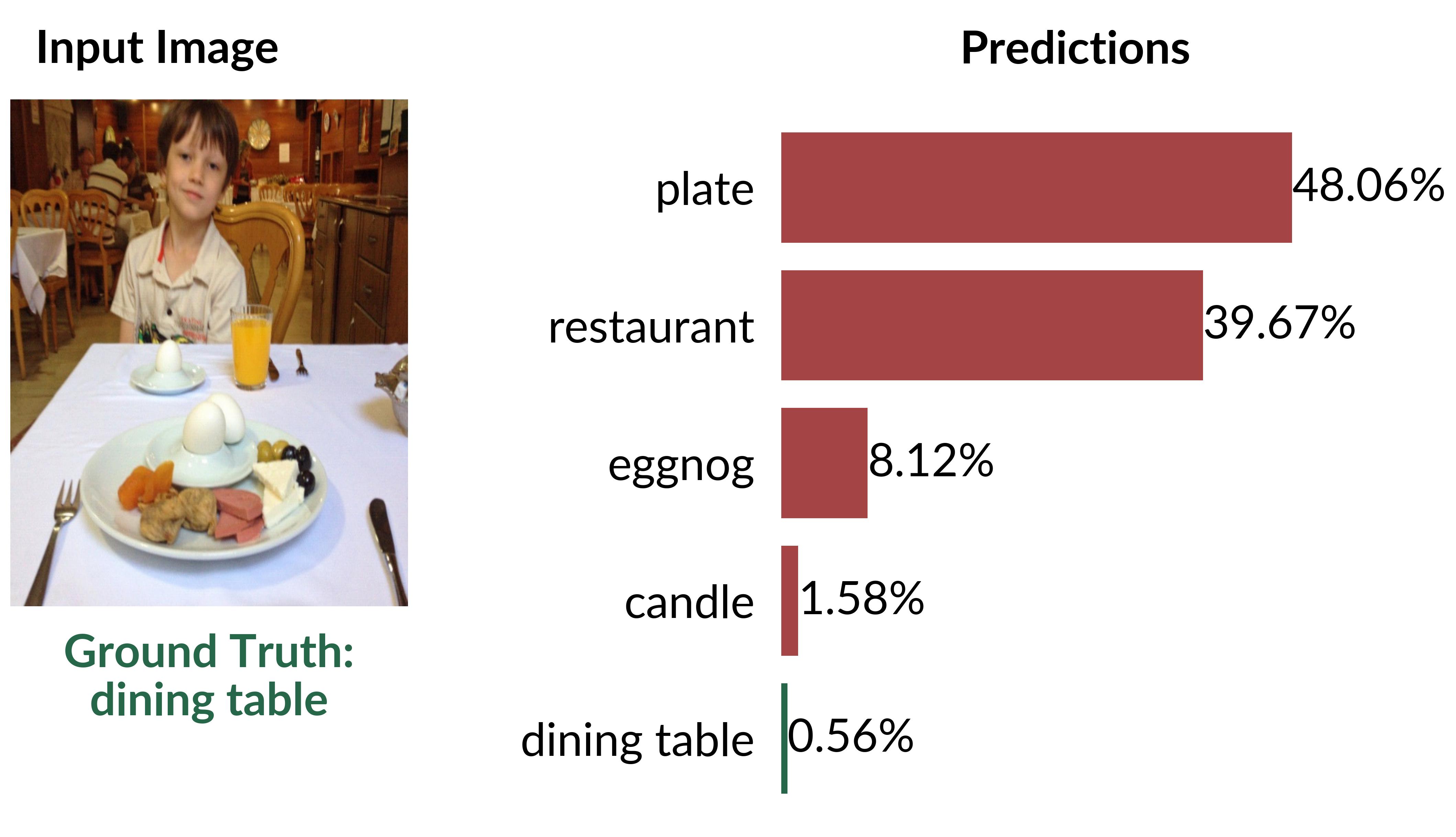}
    \end{minipage}

    \vspace{1em}

    \begin{minipage}[b]{0.3\textwidth}
        \includegraphics[width=\textwidth]{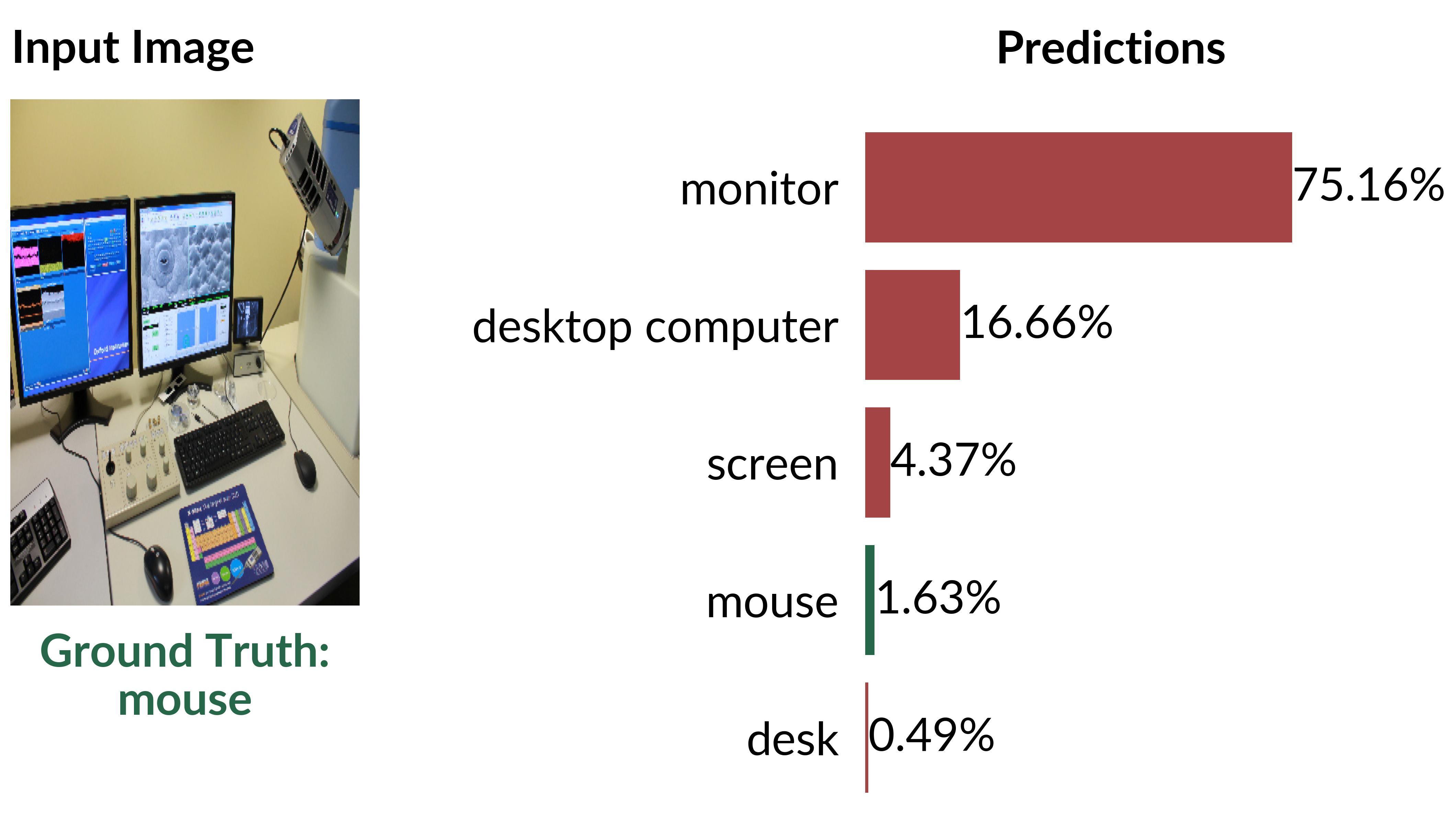}
    \end{minipage}
    \hfill 
    \begin{minipage}[b]{0.3\textwidth}
        \includegraphics[width=\textwidth]{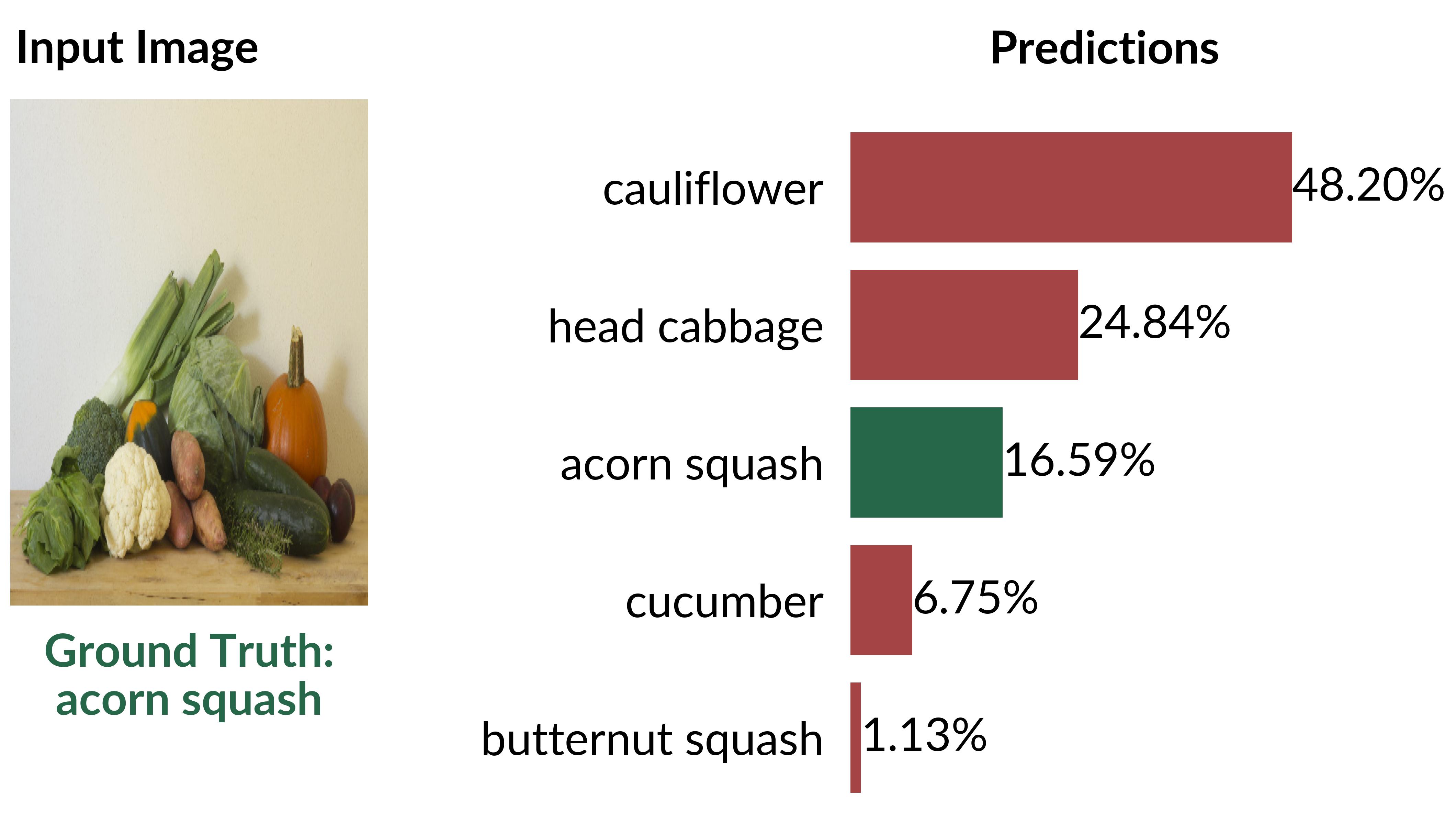}
    \end{minipage}
    \hfill 
    \begin{minipage}[b]{0.3\textwidth}
       \includegraphics[width=\textwidth]{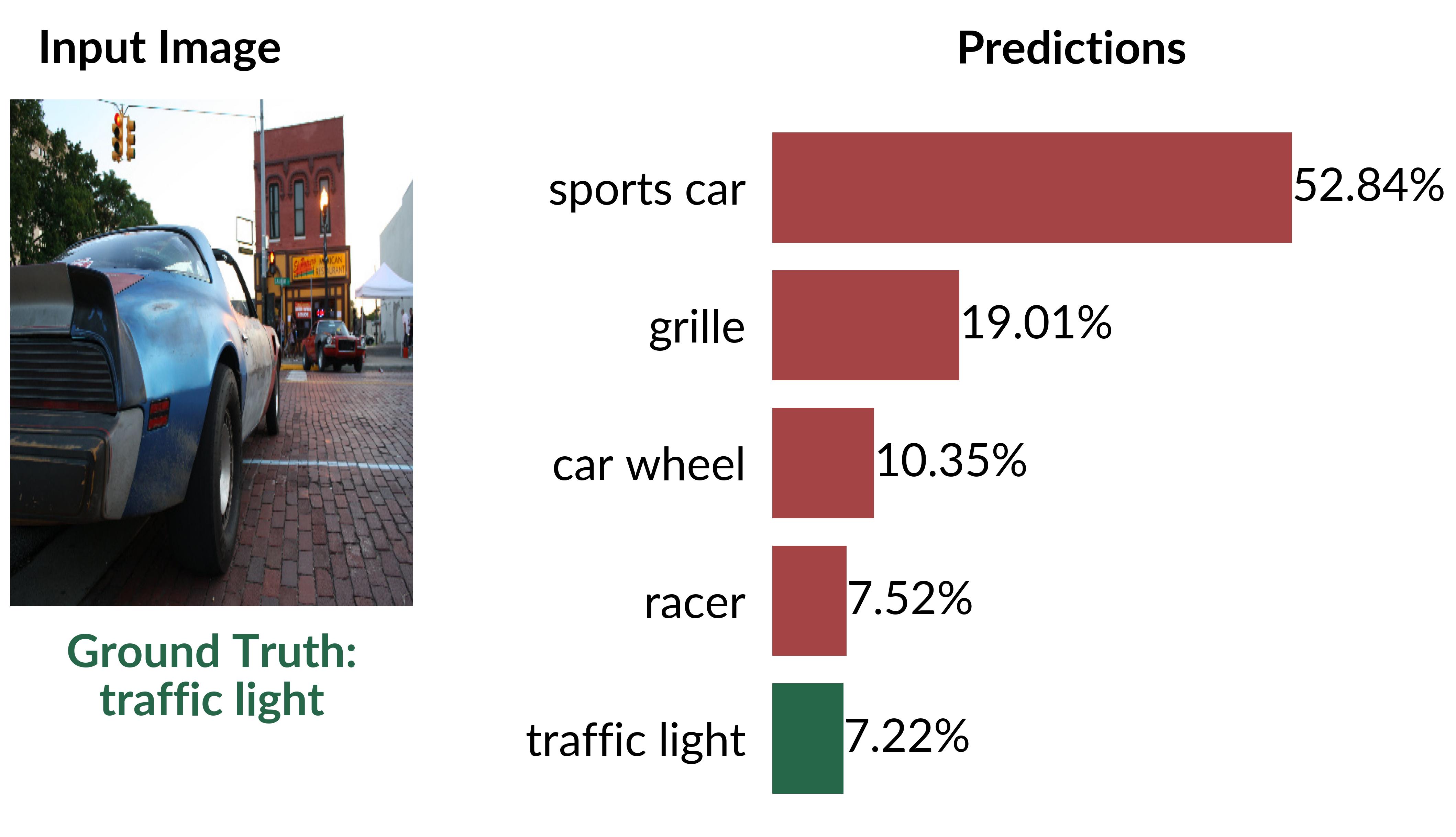}
    \end{minipage}
    \caption{Examples from ImageNetV2 showing the top-5 predictions of a pre-trained DNN model. The percentages indicate softmax scores. Ground-truth labels and correct predictions are in green; incorrect predictions are in red. While model predictions capture image complexities, evaluating models based solely on a single ground-truth label may obscure multi-label characteristics and underestimate effectiveness.}
    \label{fig:model_predictions_vs_ground_truth}
\end{figure*}

Benchmark datasets such as ImageNet~\citep{deng_imagenet_2009, russakovsky_imagenet_2015} have been instrumental in the development of deep neural networks (DNNs) for image recognition. By providing large-scale, standardized testbeds, ImageNet has catalyzed progress across supervised, self-supervised, and transfer learning techniques, enabling reproducible evaluation and fair model comparison. However, as models achieve increasingly high accuracy and move closer to real-world deployment, concerns have grown about whether these benchmarks faithfully capture the semantic and structural properties of the tasks for which they serve as proxies~\citep{barbu_objectnet_2019, recht_imagenet_2019, beyer_are_2020, shankar_evaluating_2020, tsipras_imagenet_2020, vasudevan_when_2022}.

A particularly illustrative case is ImageNetV2~\citep{recht_imagenet_2019}, a replication test set introduced in $2019$, several years after the release of ImageNet-1K~\citep{russakovsky_imagenet_2015}, the 1,000-class version of ImageNet used for the ILSVRC benchmark.\footnote{The ImageNet-1K dataset was first released for the ILSVRC challenge in 2012, but the standard reference paper describing the dataset and challenge was published in 2015~\citep{russakovsky_imagenet_2015}.}
To preserve alignment with the ImageNet-1K benchmark, ImageNetV2 was curated by the original authors using the same source (Flickr), similar upload-time filters, and identical annotation protocols. For clarity, we refer to the validation set of ImageNet-1K as \emph{ImageNetV1} throughout this paper. Most notably, the authors matched the Selection Frequency distribution from ImageNetV1 to preserve consistency in semantic difficulty and inclusion criteria; Selection Frequency refers to the proportion of Mechanical Turk annotators who affirmed each image-label pair. Despite these controls, DNNs experience a consistent $11$–$14\%$ drop in top-1 accuracy when evaluated on ImageNetV2~\citep{recht_imagenet_2019}, raising concerns about what this unexpected degradation reveals about model robustness and benchmark validity.

Prior investigations have documented numerous issues that can confound model evaluation on ImageNet and its variants. These include duplicate and near-duplicate images~\citep{vasudevan_when_2022, kisel_flaws_2024}, ambiguous or noisy labels~\citep{northcutt_pervasive_2021, luccioni_bugs_2023}, and biases related to pose, lighting, or occlusion~\citep{barbu_objectnet_2019, idrissi_imagenet-x_2022}. Of particular relevance is the growing evidence that many images contain multiple plausible object categories~\citep{beyer_are_2020, shankar_evaluating_2020, tsipras_imagenet_2020, peychev_automated_2023, kisel_flaws_2024}, yet evaluation remains constrained to a single ground-truth label. Efforts to address this have included re-annotating images with expanded label sets~\citep{beyer_are_2020, anzaku_leveraging_2024} and conducting human validation experiments to assess prediction plausibility~\citep{stock_convnets_2018, tsipras_imagenet_2020, vasudevan_when_2022}. These studies highlight how conventional top-1 metrics can underestimate model capability by penalizing predictions that are semantically correct but different from the designated ground-truth labels. To further illustrate these concerns, Figure~\ref{fig:model_predictions_vs_ground_truth} presents example top-$5$ predictions from an ImageNet pre-trained DNN. Many predictions considered incorrect under the top-$1$ metric are plausible labels for the corresponding images. 

Our work revisits the substantial drop in top-1 accuracy observed on ImageNetV2 by focusing on one of its least quantified causes: the disconnect between the multi-label nature of many images and the single-label metric used to evaluate them. While earlier studies have speculated that this mismatch may contribute to the observed degradation~\citep{shankar_evaluating_2020, anzaku_leveraging_2024}, no prior work has systematically quantified its effect across the entire ImageNetV1 and ImageNetV2 datasets or across a diverse pool of models. We address this gap through a large-scale analysis that evaluates whether multi-label-aware metrics, designed to accommodate the presence of multiple valid predictions, offer a more faithful assessment of DNN performance. Our findings provide new insights into the ImageNetV2 degradation and call for a reassessment of current evaluation practices in light of the multi-label realities of visual recognition.

\subsection{Contributions}

This study presents a structured investigation into how single-label evaluation biases distort the assessments of model generalization, particularly in the context of ImageNetV2 benchmarking. The main contributions of this study are:

\begin{itemize}
    \item \textbf{Reframing the ImageNetV2 accuracy drop as an evaluation artifact.}  
    We demonstrate that a substantial portion of the widely reported accuracy degradation on ImageNetV2 is not a model generalization failure but rather a consequence of evaluation mismatches. Conventional single-label metrics penalize models for predicting valid alternative labels, leading to misleading conclusions about DNN robustness on ImageNetV2. Our analysis shows that when multi-label characteristics are accounted for, the observed performance gap between ImageNetV1 and ImageNetV2 is substantially reduced.  

    \item \textbf{A structured evaluation framework for quantifying multi-label prediction capability (MLPC).}  
    While prior work has explored multi-label evaluation for ImageNet, we introduce a principled framework to assess how single-label-trained models handle multiple valid labels. Leveraging variable top-$k$ evaluation and average subgroup multi-label accuracy (ASMA), our method captures both whether a model predicts correct labels, and how effectively it ranks them among its top outputs. This framework enables a more nuanced quantification of MLPC in ImageNet-1K pre-trained DNNs and reveals that the effectiveness gap between ImageNetV1 and ImageNetV2 is substantially narrower than what top-1 accuracy alone suggests.

    \item \textbf{PatchML as a diagnostic tool for multi-label recognition.}  
    To further validate our findings, we introduce PatchML, a controlled multi-label dataset that disentangles multi-label recognition from dataset co-occurrence biases. Unlike real-world datasets where label co-occurrence is driven by natural scene dependencies, PatchML creates artificial compositions of objects, allowing for a direct assessment of whether models truly recognize multiple objects or rely on context. While PatchML does not reflect natural multi-label distributions, it serves as a diagnostic tool to distinguish models that exhibit robust object-level recognition from those that rely on dataset priors.  

    \item \textbf{Exposing Systematic Benchmarking Biases from Single-Label Assumptions.}
    This study exposes a critical limitation in current evaluation practices: by enforcing single-label assumptions on datasets that are inherently multi-label, standard metrics under-represent model capabilities and distort reliability assessments. As a result, models that exhibit desirable recognition behavior may be unfairly penalized, redirecting research efforts away from genuine trustworthiness challenges. Our findings emphasize the importance of adopting benchmarking protocols that reflect the complexity of real-world tasks and more accurately reveal the properties of DNNs.
\end{itemize}

\section{Related Work}
\label{sec:related_work}

\paragraph{\textbf{Explaining the Accuracy Drop of ImageNet Models on ImageNetV2}}
The ImageNetV2 dataset was introduced by \citet{recht_imagenet_2019} to replicate the ImageNetV1 validation set using the same data collection methodology. However, models pre-trained on ImageNet-1K exhibited a top-1 accuracy drop of $11$–$14\%$ when evaluated on ImageNetV2. \citet{recht_imagenet_2019} attributed this decline to ImageNetV2 images being slightly harder, while \citet{engstrom_identifying_2020} proposed statistical bias in the dataset replication process as a contributing factor. Their analyses suggested that after accounting for the statistical bias, only $3.6\%$ of the accuracy gap remained unexplained. \citet{anzaku_re-assessing_2025} further argued that reliance on top-1 accuracy alone, without leveraging model uncertainty, can amplify the perceived accuracy gap. Despite these efforts, none of these studies systematically examined the role of multi-label images in ImageNetV2, or identified dataset factors that are responsible for the gap.

Our work builds upon these prior findings but reframes the discussion by establishing the multi-label discrepancy as a primary explanatory factor. Through systematic quantification of multi-label prevalence in ImageNetV1 and ImageNetV2, we demonstrate through extensive experiments that a substantial portion of the observed accuracy degradation in ImageNetV2 arises from the higher incidence of multi-label images rather than genuine model failures. This insight challenges conventional interpretations of generalization performance, revealing that many models possess strong multi-label recognition capabilities that are not fully captured under single-label evaluation.

\paragraph{\textbf{Attributing ImageNetV2 Accuracy Degradation to Multi-Label Prevalence}}

While earlier works recognized that many ImageNet images contain multiple valid object categories, only recent studies have formalized this through refined multi-label annotations. \citet{beyer_are_2020} introduced ReaL, a multi-label version of ImageNetV1, showing that many top-1 errors under conventional evaluation corresponded to valid but unannotated labels. \citet{anzaku_leveraging_2024} extended this effort to ImageNetV2, producing refined multi-label annotations for the full dataset. \citet{shankar_evaluating_2020} independently estimated that $30.0\%$ of ImageNetV1 and $34.4\%$ of ImageNetV2 images contain multiple valid labels, based on a manually labeled sample of $1,000$ images from each dataset, although the specific ImageNetV2 variant was not specified. While these works improved label quality, neither systematically investigated how multi-label prevalence affects the observed accuracy drop on ImageNetV2.

\paragraph{\textbf{ImageNet Benchmark Underestimating DNN Effectiveness}}

Several studies have challenged the adequacy of single-label evaluation metrics in ImageNet benchmarking. Human evaluator reviews have shown that many model predictions previously marked as errors were, in fact, correct identifications of valid but unannotated objects~\citep{stock_convnets_2018, tsipras_imagenet_2020, vasudevan_when_2022}. Most recently, \citet{vasudevan_when_2022} reported that nearly half of the errors made by top-performing ImageNet classifiers fell into this category. \citet{taesiri_imagenet-hard_2023} further illustrated that models could achieve high apparent accuracy by strategically ``zooming in" on the most discriminative image regions, suggesting that many apparent errors arise from dataset annotation constraints and evaluation methodology rather than true model limitations.

Our study aligns with prior work in arguing that single-label benchmarks underestimate model capabilities. We advance this argument by demonstrating that such underestimation has specific and measurable consequences, notably contributing to the apparent accuracy drop observed in ImageNetV2. Through multi-label-aware evaluation, we show that models retain strong predictive abilities on ImageNetV2, despite being penalized under strict single-label evaluation metrics.

\paragraph{\textbf{Multi-label Learning in ImageNet}}
While ImageNet models are typically trained under a single-label assumption, the dataset itself contains many images with multiple valid labels. Even if providing full multi-label annotations for training remains challenging, test sets must reflect this property to ensure reliable assessments of generalization and robustness.

To address the limitations of single-label supervision, recent research has explored multi-label training strategies for ImageNet. Single Positive Multi-label Learning (SPML) \citep{cole_multi-label_2021} infers additional labels from a single known positive label per image, enabling multi-label training without exhaustive annotations. \citet{verelst_spatial_2023} demonstrated that SPML scales effectively to large datasets, including ImageNet-1K, while reducing annotation costs. Another approach, ReLabel~\citep{yun_re-labeling_2021}, refines label assignments to improve consistency in localized object recognition.

These efforts aim to incorporate the multi-label nature of ImageNet in the training stage, aligning with the broader argument that the dataset should be treated as such. Our focus, however, is on the consequences of disregarding this property during evaluation, showing that single-label benchmarks misrepresent the capabilities of DNNs pre-trained on ImageNet.

\section{Methodology}

The main objective of this study is to investigate how the assumption that each image has a single correct label impacts the reported accuracy drop on ImageNetV2 for ImageNet pre-trained DNNs, even though these models are not inherently designed to produce multi-label predictions. This section outlines the methodology used to derive multi-label predictions from these models and to evaluate them. The methodology is not intended for deployment; rather, it serves as a diagnostic tool to assess how enforcing a single-label evaluation affects accuracy measurements on datasets where individual images may contain multiple valid labels.

Section~\ref{subsec:variable_topk} presents the variable top-$k$ method used to derive multi-label predictions from multi-class classification models. Section~\ref{subsec:patchml_dataset_generation} describes how we generated the PatchML dataset, while Section~\ref{subsec:mlpc_metrics} details the multi-label evaluation metrics used in this study.

\subsection{Variable Top-$k$ Multi-label Predictions}
\label{subsec:variable_topk}

Conventional multi-class classification models with softmax outputs are limited by their reliance on a single top-1 prediction, making them unsuitable for datasets where instances belong to multiple classes. Existing evaluation methods, such as top-$5$ accuracy and ReaL accuracy, attempt to assess the MLPC of ImageNet pre-trained models but have distinct limitations. Top-5 accuracy verifies whether at least one of the five highest-ranked predictions is correct but does not evaluate whether all relevant categories are identified. ReaL accuracy expands the ground-truth label set, but considers only the top-ranked prediction, overlooking cases where multiple valid predictions exist. To address these shortcomings, we propose a variable top-\( k \) selection mechanism, where \( k \) is derived from the number of valid ground-truth labels for each image. This approach allows us to generate a multi-label prediction set to evaluate the MLPC of models pre-trained on ImageNet by utilizing the metrics explained in Section~\ref{subsec:mlpc_metrics}.

Given a set of test datapoints \( \mathbf{X} = \{ \mathbf{x}_1, \mathbf{x}_2, \ldots, \mathbf{x}_N \} \) and their corresponding softmax outputs \( \hat{\mathbf{Y}} = \{ \hat{\mathbf{y}}_1, \hat{\mathbf{y}}_2, \ldots, \hat{\mathbf{y}}_N \} \), with \( \hat{\mathbf{y}}_i \in \mathbb{R}^C \) representing the predicted probability distribution over \( C \) classes for the \( i \)-th datapoint, we define \( k_i \) as the number of ground-truth labels for the \( i \)-th datapoint. For each datapoint \( \mathbf{x}_i \), the top-\( k_i \) predictions are obtained by selecting the indices corresponding to the highest \( k_i \) values in \( \hat{\mathbf{y}}_i \).

\subsection{Patch-based Multi-Label Dataset Generation}
\label{subsec:patchml_dataset_generation}

\begin{figure*}[!t]
    \centering
    \includegraphics[scale=0.27]{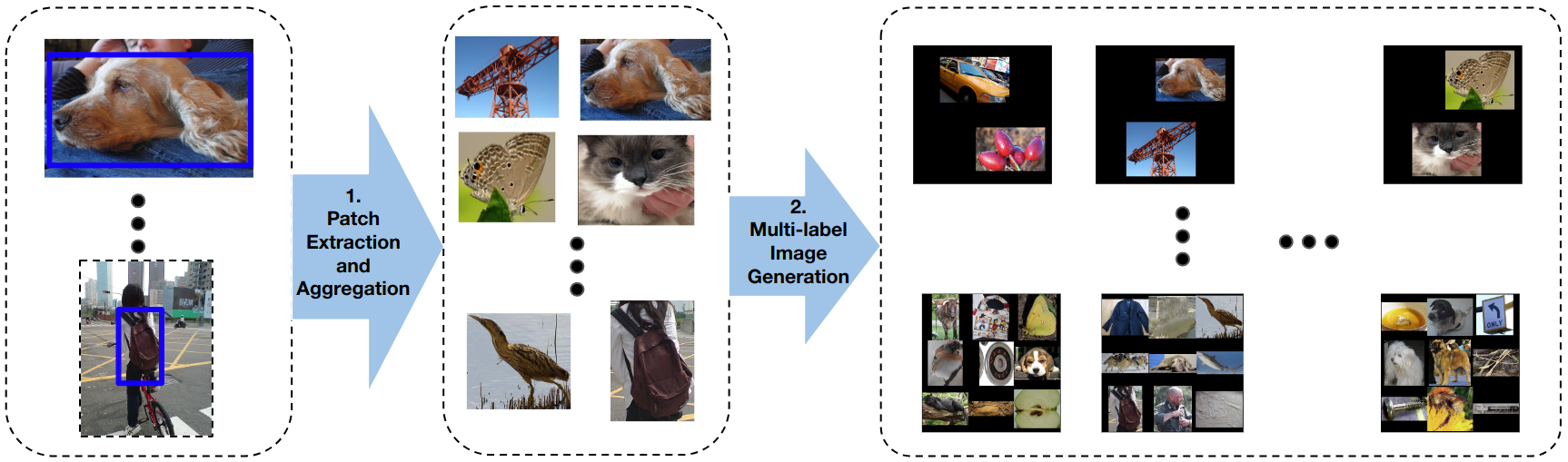}
    \caption{An illustration of the PatchML dataset creation process, organized into two main stages: (1) Patch Extraction and Aggregation, where object regions are cropped and pooled; and (2) Multi-label Image Generation, where a predefined number of patches are randomly sampled without replacement and placed on blank backgrounds to create new multi-label images with corresponding ground-truth label sets.}
    \label{fig:patchML_overview}
\end{figure*}

\paragraph{\textbf{Motivation}}
Existing multi-label annotations for ImageNetV1 and ImageNetV2 are valuable but inherently reflect real-world semantic relationships between co-occurring objects. As a result, it remains unclear whether models genuinely recognize multiple objects independently or primarily rely on learned contextual cues. A rigorous assessment of the MLPC of DNNs requires isolating object recognition from these contextual relationships. To achieve this, we introduce a synthetic multi-label dataset, PatchML, specifically constructed to control and minimize semantic co-occurrence biases. PatchML thus provides a complementary diagnostic tool to effectively evaluate the intrinsic multi-label capabilities of pre-trained models. 

\paragraph{\textbf{Dataset Construction}}
PatchML is generated in two stages: \emph{Patch Extraction and Aggregation} followed by \emph{Multi-label Image Generation}. In the first stage, object patches are extracted from annotated ImageNet images, forming a labeled pool of patches denoted by $\mathcal{S}$. In the second stage, we construct composite images by randomly sampling patches without replacement from this patch pool. Specifically, for each predefined patch count \(k \in \mathcal{K} = \{2, 3, 4, 6, 9\}\), and corresponding patch dimensions \(p \in \mathcal{P} = \{256, 256, 256, 170, 128\}\), each selected patch is resized proportionally to a square of size \(p\). These resized patches are then placed randomly within separate, non-overlapping grid cells on a fixed-size black canvas of dimensions \(F \times F\), where \(F = 512\). Small random offsets within grid cells introduce spatial diversity. The labels for each composite image are assigned by taking the union of labels from its constituent patches. Algorithm~\ref{algo:patchml_dataset_generation} provides detailed algorithmic steps, and the visual illustration of the PatchML dataset creation process is presented in Figure~\ref{fig:patchML_overview}.

\begin{algorithm*}[!thpb]
\caption{Patch-based Multi-Label (PatchML) Dataset Generation}
\label{algo:patchml_dataset_generation}
\begin{algorithmic}[1]

\Procedure{GeneratePatchMLDataset}{$\mathcal{S}, \mathcal{K}, \mathcal{P}, F$}
\footnotesize
    \State $\mathcal{D} \gets \emptyset$ \Comment{Initialize dataset}
    \For{each $(k, p) \in \text{zip}(\mathcal{K}, \mathcal{P})$}
        \State $c \gets \lfloor F/p \rfloor$ \Comment{Grid dimension (columns and rows)}
        \While{$|\mathcal{S}| \geq k$}
            \State $I \gets \text{zeros}(F, F)$ \Comment{Initialize empty black canvas}
            \State $\mathcal{S}_k \gets \text{random\_sample}(\mathcal{S}, k)$ \Comment{Sample patches without replacement}
            \State $\mathcal{U} \gets \emptyset$ \Comment{Occupied grid cells}
            \State $\mathcal{L} \gets \emptyset$ \Comment{Set of labels}
            
            \For{$i \gets 0$ \textbf{to} $k-1$}  
                \State $\text{row} \gets \lfloor i/c \rfloor$, $\text{col} \gets i \bmod c$
                \State $(I, \text{success}, \mathcal{U}) \gets$ \Call{PlacePatchInGrid}{$I, \mathcal{S}_k[i], \text{row}, \text{col}, p, F, \mathcal{U}$}
                \If{success}
                    \State $\mathcal{L} \gets \mathcal{L} \cup \{\text{label}(\mathcal{S}_k[i])\}$ 
                \EndIf
            \EndFor
            \State $\mathcal{D}[\text{id}(I)] \gets \mathcal{L}$ \Comment{Store composite image and labels}
            \State $\mathcal{S} \gets \mathcal{S} \setminus \mathcal{S}_k$ \Comment{Remove used patches}
        \EndWhile
    \EndFor
    \State \textbf{return} $\mathcal{D}$
\EndProcedure
\Statex
\hrulefill
\Procedure{PlacePatchInGrid}{$I, s, \text{row}, \text{col}, p, F, \mathcal{U}$}
    \If{$(\text{row}, \text{col}) \in \mathcal{U}$}
        \State \textbf{return} $(I, \text{false}, \mathcal{U})$ \Comment{Grid cell occupied}
    \EndIf
    \State $s' \gets \text{resize}(s, p)$
    \State $(w, h) \gets \text{dimensions}(s')$
    \State $x_{\text{off}} \sim \mathcal{U}(0, p - w)$, $y_{\text{off}} \sim \mathcal{U}(0, p - h)$
    \State $x \gets \text{col} \cdot p + x_{\text{off}}$,\quad$y \gets \text{row} \cdot p + y_{\text{off}}$
               
    \If{$x + w \leq F$ and $y + h \leq F$}
        \State $I[y:y+h, x:x+w] \gets s'$ \Comment{Place resized patch on canvas}
        \State $\mathcal{U} \gets \mathcal{U} \cup \{(\text{row}, \text{col})\}$
        \State \textbf{return} $(I, \text{true}, \mathcal{U})$
    \EndIf
    \State \textbf{return} $(I, \text{false}, \mathcal{U})$ \Comment{Patch placement invalid}
\EndProcedure
\end{algorithmic}
\end{algorithm*}

\paragraph{\textbf{Intended Utility}}
PatchML serves explicitly as a diagnostic tool rather than as a dataset for model training. Unlike patch-based augmentation techniques such as CutMix~\citep{yun_cutmix_2019} and RICAP~\citep{takahashi_ricap_2018}, which preserve contextual relationships to improve generalization during training, PatchML completely removes these contextual dependencies. By controlling object composition, PatchML enables a precise evaluation of the intrinsic multi-label recognition capability of a model, independent of any learned co-occurrence biases. This diagnostic approach is particularly valuable for investigating anomalies in performance, such as the accuracy drop observed on ImageNetV2, clarifying whether these discrepancies reflect genuine model weaknesses or artifacts introduced by evaluation constraints.

\paragraph{\textbf{Domain Shift Considerations}}
PatchML intentionally simplifies natural image distributions by isolating objects onto plain black backgrounds, entirely removing realistic scene complexity such as backgrounds, occlusions, or object interactions. This deliberate simplification ensures clarity on valid object labels, preventing ambiguity from unlabeled background objects or context-driven label associations. Although such simplification introduces a clear distribution shift relative to real-world datasets, it precisely supports the diagnostic objective of PatchML: enabling a focused evaluation of object-level recognition independently from contextual influences. This controlled approach aligns with established scientific practices that isolate specific model behaviors for rigorous assessment, providing clearer insights into multi-label recognition capability.

\subsection{Model Assessment Methodology}
\label{subsec:mlpc_metrics}

After obtaining multi-label predictions from DNNs pre-trained on ImageNet, we employed three key metrics to assess their MLPC: top-1 accuracy, ReaL accuracy, and average subgroup multi-label accuracy (ASMA). These metrics were chosen to provide a comprehensive evaluation while maintaining consistency with conventional classification assessment approaches.

\paragraph{\textbf{Top-1 Accuracy}}
The top-1 accuracy metric, widely used in multi-class classification, measures the proportion of datapoints for which the highest predicted label of a model matches the single ground-truth label. Formally, for each datapoint $\mathbf{x}_i$, with predicted label $\hat{y}_i$ and ground-truth label $y_i^{\text{gt}}$, top-1 accuracy is defined as:
$\frac{1}{N} \sum_{i=1}^{N} \mathbb{I}(\hat{y}_i = y_i^{\text{gt}})$,
where $\mathbb{I}(\cdot)$ is the indicator function, and $N$ is the total number of datapoints. This formulation aligns with traditional definitions of accuracy used in multi-class settings, such as those discussed by~\citet{deng_deep_2020} in the context of ImageNet evaluation.

\paragraph{\textbf{ReaL Accuracy}}
The ReaL accuracy extends top-1 accuracy by accepting matches with any label from an expanded set of plausible ground-truth labels. This metric, introduced by~\citet{beyer_are_2020}, better handles cases where multiple valid labels exist. Formally, for a datapoint $\mathbf{x}_i$ with predicted label $\hat{y}_i$ and plausible label set $\mathbf{y}_i^{\text{plaus}}$, the ReaL accuracy is calculated as:
$\frac{1}{N} \sum_{i=1}^{N} \mathbb{I}(\hat{y}_i \in \mathbf{y}_i^{\text{plaus}})$.

\paragraph{\textbf{Average Subgroup Multi-Label Accuracy}}

To address the imbalance in the distribution of label counts in multi-label datasets, we introduce ASMA, extending the example-based accuracy metric from multi-label learning~\citep{zhang_review_2014}.

We define subgroups based on the number of valid labels for each image. Each subgroup \(g\) contains all datapoints with exactly \(g\) ground-truth labels, where \(g\) ranges from 0 to the maximum number of labels observed in any datapoint. For example, a subgroup \(g=3\) consists of all images that have exactly three valid ground-truth labels.

For a dataset with \(C\) classes, let datapoint \(\mathbf{x}_{g,i}\) in subgroup \(g\) have ground-truth labels \(\mathbf{y}_{g,i}^{\text{gt}} \in \{0,1\}^C\) and predictions \(\hat{\mathbf{y}}_{g,i} \in \{0,1\}^C\), where \(i \in \{1,\ldots,N_g\}\) and \(N_g\) is the number of datapoints in subgroup \(g\).

The label-wise accuracy for a datapoint is defined as:
\[
\frac{1}{C} \sum_{c=1}^{C} \mathbb{I}(y_{g,i,c}^{\text{gt}} = \hat{y}_{g,i,c}),
\]
where \(\mathbb{I}(\cdot)\) is the indicator function.

The subgroup accuracy \(A_g\) is the average label-wise accuracy across all datapoints in subgroup \(g\):
\[
A_g = \frac{1}{N_g} \sum_{i=1}^{N_g} \frac{1}{C} \sum_{c=1}^{C} \mathbb{I}(y_{g,i,c}^{\text{gt}} = \hat{y}_{g,i,c}).
\]

The ASMA is then computed as the mean of all subgroup accuracies:
\[
\text{ASMA} = \frac{1}{G} \sum_{g=0}^{G-1} A_g,
\]
where \(G\) is the total number of subgroups, corresponding to the maximum observed label count plus one.


\section{Experiments}
\label{sec:experiments_and_results}
\subsection{Experimental Setup}
\label{subsec:experiment_setup}

\paragraph{\textbf{Test Dataset Description}}

\begin{table}[!thbp]
\centering
\caption{The number of images per label count for the ImageNetV1 and ImageNetV2 datasets. For ImageNetV1, we used the ReaL labels~\citep{beyer_are_2020}, which provided refined annotations for $46,837$ out of the $50,000$ available images in the ImageNetV1 dataset. For ImageNetV2, we used the annotations created by~\citet{anzaku_leveraging_2024}, totaling $9,858$ images from the ImageNetV2 dataset.}
\label{table:label_count_imagenet_dsets}
\begin{tabular}{l*{6}{c}}
\toprule
\multirow{2}{*}{\textbf{Dataset}} & \multicolumn{6}{c}{\textbf{Label Count}} \\ \cmidrule(l){2-7}
 & \textbf{1} & \textbf{2} & \textbf{3} & \textbf{4} & \textbf{5} & \textbf{$>$5} \\ \midrule
ImageNetV1 & 39,394 & 5,408 & 1,319 & 411 & 161 & 144 \\
ImageNetV2 & 5,083 & 2,385 & 1,306 & 628 & 237 & 232 \\
\bottomrule
\end{tabular}
\end{table}

\begin{table}[!thbp]
\centering
\caption{The number of images per label count across the PatchML dataset variants. Each variant was obtained using a different seed.}
\label{table:patchml_label_counts}
\setlength{\tabcolsep}{6pt}
    \begin{tabular}{lccccc}
    \toprule
    \multirow{2}{*}{\textbf{Dataset}} & \multicolumn{5}{c}{\textbf{Label Count}} \\
    \cmidrule(l){2-6}
     & \textbf{2} & \textbf{3} & \textbf{4} & \textbf{6} & \textbf{9} \\
    \midrule
    PatchML1 & 26,778 & 17,870 & 13,281 & 8,766 & 5,736 \\
    PatchML2 & 26,790 & 17,840 & 13,306 & 8,787 & 5,739 \\
    PatchML3 & 26,777 & 17,874 & 13,282 & 8,760 & 5,718 \\
    PatchML4 & 26,793 & 17,852 & 13,296 & 8,746 & 5,721 \\
    PatchML5 & 26,786 & 17,853 & 13,296 & 8,784 & 5,731 \\
    \bottomrule
    \end{tabular}
\end{table}

We utilized three datasets to assess the pre-trained DNNs in this study. These datasets were introduced in Section~\ref{sec:related_work} and are the ImageNetV1, ImageNetV2, and PatchML test datasets. ImageNetV1 is the validation set of the ImageNet-1K dataset that comprises $50,000$ images. We used the original single-label ground-truth labels for this dataset to compute the top-1 accuracy and used the ReaL labels~\citep{beyer_are_2020} to compute the ReaL accuracy. For ImageNetV2, we used the refined labels generated by~\citet{anzaku_leveraging_2024}, which account for multi-label images. The PatchML dataset was created from ImageNetV1 and the available object detection annotations, following the methodology described in Section~\ref{subsec:patchml_dataset_generation}. The number of images of the ImageNet-related datasets across different label counts is provided in Table~\ref{table:label_count_imagenet_dsets}. Since the generation process of PatchML involves random selection and placement of cropped object patches, we created five variants using different seeds to account for variability arising from the sampling process. The resulting datasets are summarized in Table~\ref{table:patchml_label_counts}.

\paragraph{\textbf{Description of the Evaluated Models}}
We evaluate a total of $315$ DNNs that were pre-trained on ImageNet and sourced from the TIMM repository~\citep{wightman_pytorch_2019}. They were selected to ensure broad and meaningful coverage across architectural families, training techniques, and accuracy ranges. The collection spans traditional convolutional architectures such as ResNet~\citep{he_deep_2016}, MobileNet~\citep{howard_mobilenets_2017}, and EfficientNet~\citep{tan_efficientnet_2019}, as well as more recent CNN-based variants like ConvNeXt and ConvNeXt V2~\citep{woo_convnext_2023}. Vision Transformers and hybrid attention-based models are also extensively represented, including Vision Transformer (ViT)~\citep{dosovitskiy_image_2021}, Data-efficient image Transformer (DeiT)~\citep{touvron_training_2021}, BERT Pre-Training of Image Transformers (BEiT)~\citep{dong_beit_2022}, Swin Transformers~\citep{liu_swin_2021}, Class-Attention in Image Transformers (CaiT)~\citep{touvron_going_2021}, and Vision Outlooker (VOLO)~\citep{yuan_volo_2023}.

Crucially, the selection encompasses a wide range of training strategies: from standard supervised training on ImageNet-1K and the full ImageNet, with approximately $22,000$ classes, to self-supervised and semi-supervised approaches leveraging masked image modeling (MIM). These include methods such as Fully Convolutional Masked Autoencoders (FCMAE)~\citep{woo_convnext_2023}, MIM with untokenized vision-text supervision~\citep{sun_eva-clip_2023}, and dense patch-level supervision models like VOLO~\citep{yuan_volo_2023}.
Several models in the benchmark were distilled from teacher networks originally pre-trained on massive weakly supervised datasets like Instagram-1B~\citep{mahajan_exploring_2018}, indirectly introducing large-scale external knowledge through techniques exemplified in SWAG's RegNetY-16GF framework~\citep{singh_revisiting_2022}.

To ensure comprehensive coverage, we include the top-$100$ models by ImageNet-1K top-1 accuracy and supplement them with 215 additional models randomly sampled across the full range of TIMM models. This sampling was stratified to capture diversity in architecture type, parameter scale, training strategy, and performance level -- from lightweight CNNs and early ViT variants to recent large-scale models that push the state-of-the-art. This approach avoids performance bias and ensures that our analysis spans the full design space of modern DNNs, from historical developments like ResNet to cutting-edge architectures like ConvNeXt V2, and from mobile-optimized models to compute-intensive ones. It also covers a wide spectrum of supervision paradigms, ranging from clean labeled data to web-scale weak supervision. The complete list of models is provided in the code repository accompanying this paper.

\subsection{Results and Discussion}

\subsubsection{Prevalence of Multi-label Images}
\begin{figure}[!thbp]
    \centering
    \includegraphics[width=0.75\linewidth]{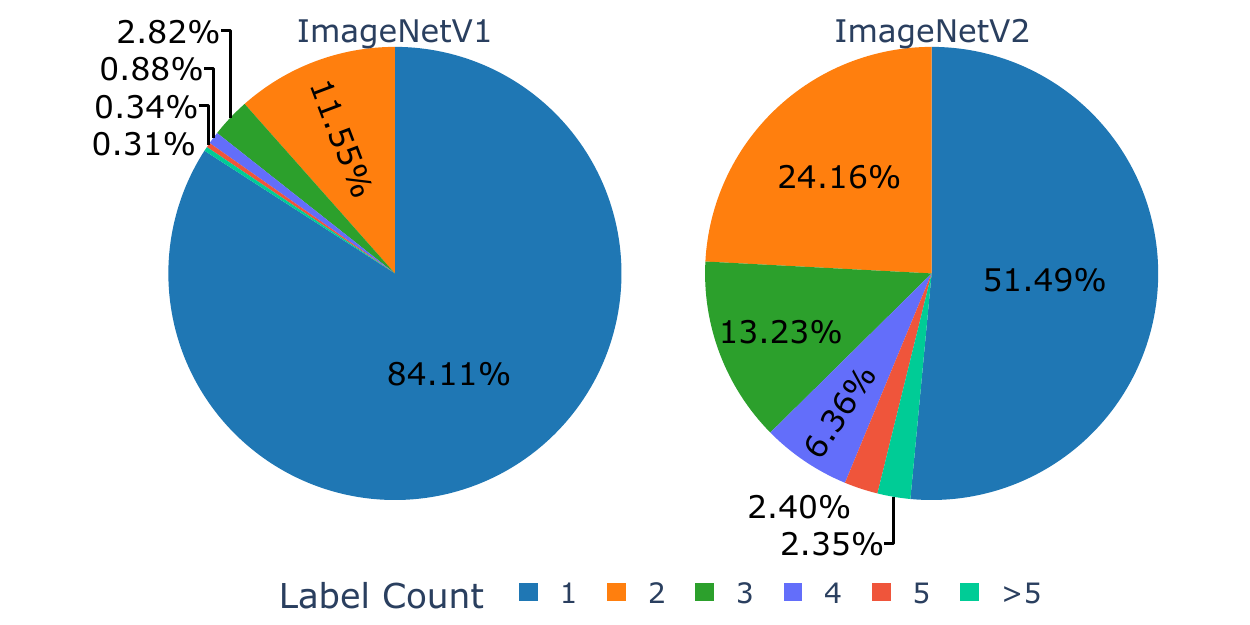}
    \caption{Distribution of images by number of ground-truth labels in ImageNetV1 and ImageNetV2, with images having more than five labels grouped as ``$>5$''.}
    \label{fig:pie_chart_multi_label_proportions}
\end{figure}
We begin our analysis by quantifying the proportion of multi-label images in ImageNetV1 and ImageNetV2. To do this, we utilize the re-assessed labels from ReaL~\citep{beyer_are_2020} for ImageNetV1 and the refined labels introduced by~\citet{anzaku_leveraging_2024} for ImageNetV2. As visualized in Figure~\ref{fig:pie_chart_multi_label_proportions}, ImageNetV2 contains a substantially larger proportion of images with more than one valid label (approximately $48\%$) compared to ImageNetV1 (around $16\%$).

These estimates are consistent with prior findings, although they differ in magnitude. For instance, \citet{shankar_evaluating_2020} reported multi-label rates of $30.0\%$ for ImageNetV1 and $34.4\%$ for ImageNetV2 based on $1,000$ randomly sampled images. However, there are two key differences between their analysis and ours. First, we used the full dataset annotations instead of a randomly selected subset. Second, the ImageNetV2 variant used by~\cite{shankar_evaluating_2020} is unclear; we utilized the MatchedFrequency ImageNetV2 variant, which is the most faithful reconstruction of the original ImageNetV1 collection process. These differences allow for a more comprehensive estimation of multi-label prevalence in ImageNetV2.

While differences in re-annotation protocols and sampling strategies may account for the variation in reported prevalence, the broader observation remains clear: both datasets include a non-negligible number of multi-label images, with ImageNetV2 containing a considerably higher proportion of multi-label images. This characteristic is not merely a dataset artifact---it has implications for how model performance should be interpreted under standard evaluation protocols. These implications are examined in the following sections.

\subsubsection{Accuracy Analysis on ImageNetV1 and ImageNetV2}

\begin{figure*}[htbp]
    \centering
    \begin{minipage}[b]{0.32\textwidth}
        \includegraphics[width=\textwidth]{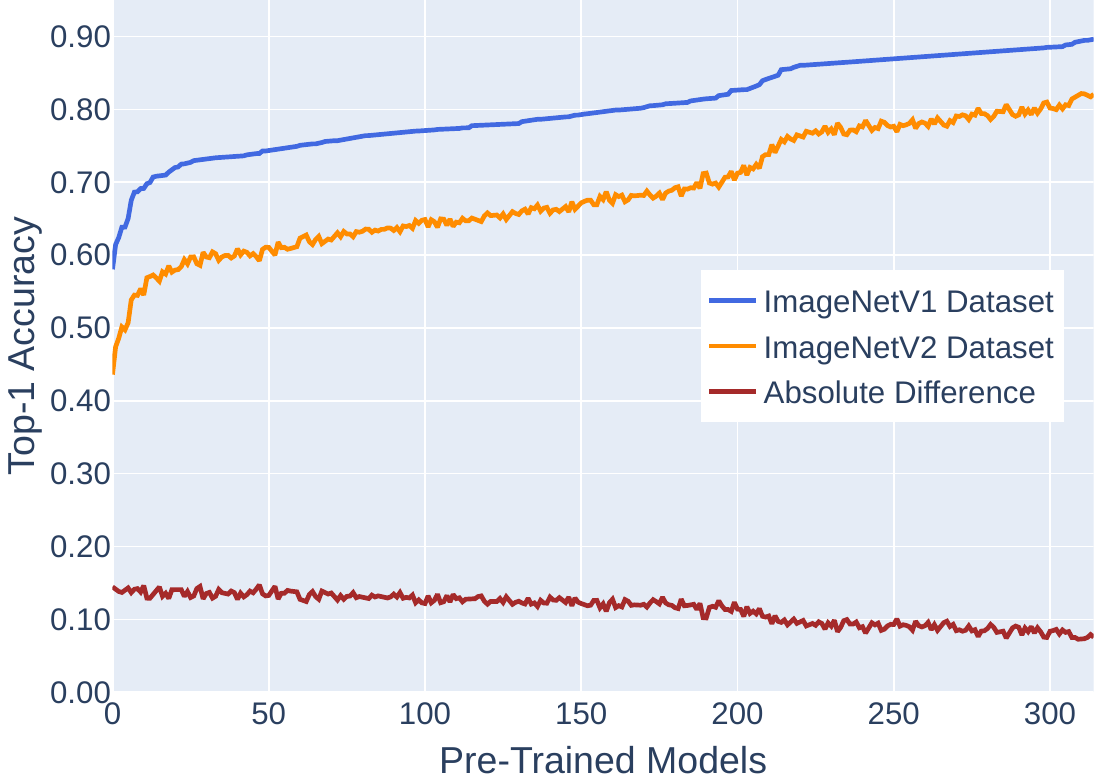}
        \subcaption{}
        \label{fig:top-1_v1_v2}
    \end{minipage}
    \begin{minipage}[b]{0.32\textwidth}
        \includegraphics[width=\textwidth]{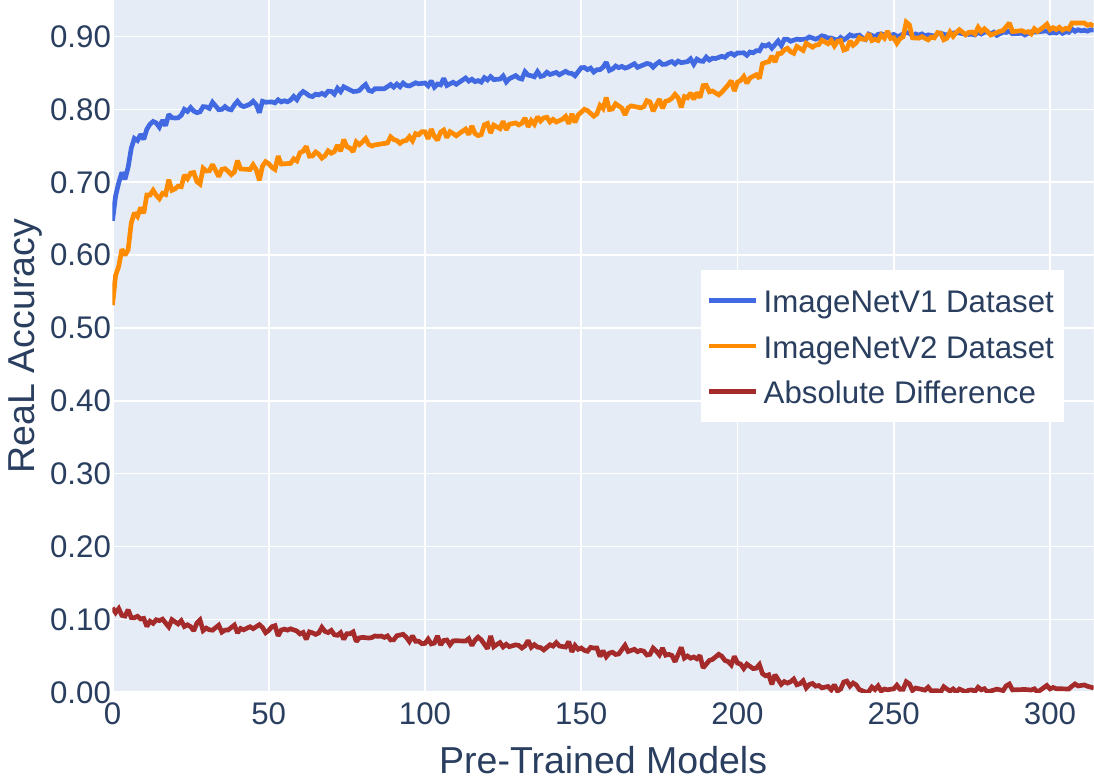}
        \subcaption{}
        \label{fig:real_v1_v2}
    \end{minipage}
    \begin{minipage}[b]{0.32\textwidth}
        \includegraphics[width=\textwidth]{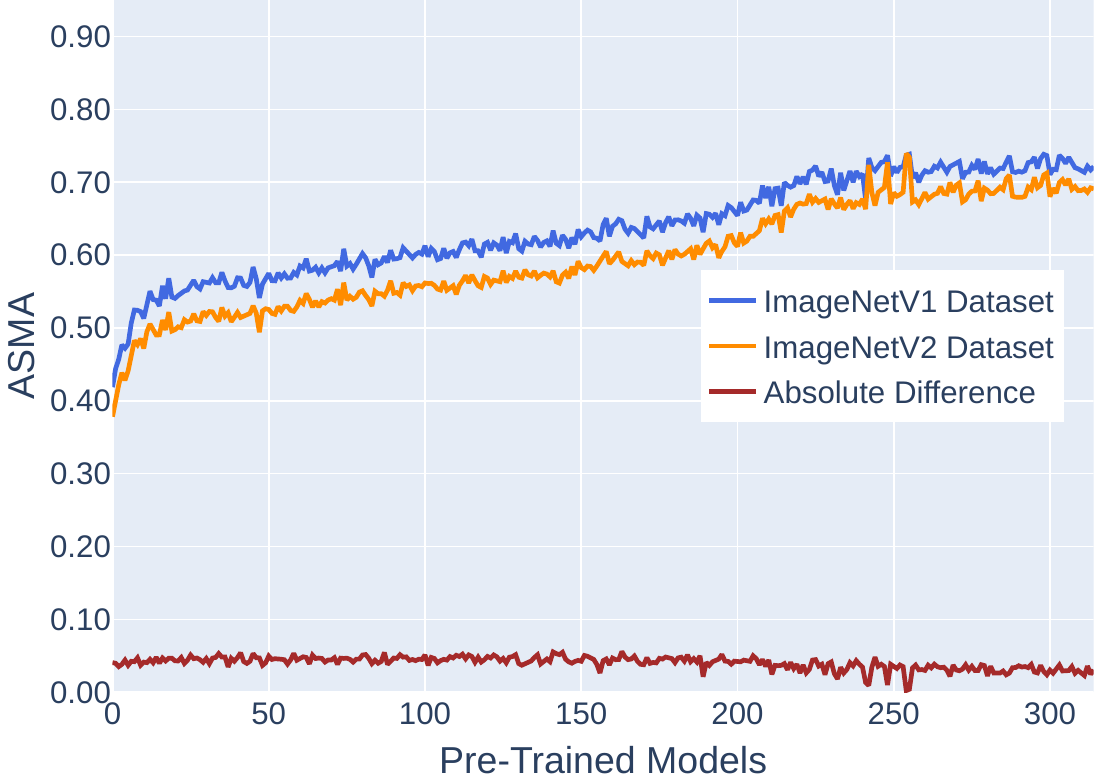}
        \subcaption{}
        \label{fig:asma_v1_v2}
    \end{minipage}

   \caption{Accuracies obtained by assessing $315$ pre-trained DNNs on \textit{ImageNetV1} and \textit{ImageNetV2}: (a) Top-1 accuracy, (b) ReaL accuracy, and (c) ASMA, all plotted against label count. The absolute difference represents the accuracy difference between the two datasets. Models in all plots are sorted by Top-1 accuracy on ImageNetV1.}
    \label{fig:various_mlpc_on_V1_and_V2}
\end{figure*}

\paragraph{\textbf{Top-1 Accuracy Can Overstate DNN Performance Gaps}}
In Figure~\ref{fig:top-1_v1_v2}, we present the top-1 accuracies for $315$ ImageNet-pretrained DNNs evaluated on both ImageNetV1 and ImageNetV2. Consistent with prior 
work~\citep[see Section~\ref{sec:related_work}]{recht_imagenet_2019} 
, the top-1 accuracy on ImageNetV2 is lower for each model, with drops of roughly $6\%$ to $14\%$. At first glance, these declines appear to indicate substantially weaker generalization to ImageNetV2.

However, when we use ReaL accuracy (Figure~\ref{fig:real_v1_v2}) -- which accepts any valid label as correct -- the difference between ImageNetV1 and ImageNetV2 narrows noticeably. Seventy-eight models show a gap of less than $1\%$, and the overall range shrinks to $0\%$–$11\%$. Although ReaL accuracy is more flexible than top-1 accuracy, it still only evaluates the single best-predicted label per image of a model and therefore does not evaluate whether other valid labels are identified.

To address this issue, we apply ASMA, a stricter multi-label metric that assesses how comprehensively each model captures the entire set of valid labels for a given image. ASMA normalizes results based on the number of possible labels, offering a more thorough evaluation. As shown in Figure~\ref{fig:asma_v1_v2}, the gap under ASMA diminishes further to $0\%$–$6\%$, with four models showing less than a $1\%$ decrease. These findings indicate that the commonly reported accuracy drop on ImageNetV2 is largely a consequence of evaluating models with single-label metrics that fail to account for multiple valid labels, rather than evidence of a genuine decline in model performance under distribution shift.

\paragraph{\textbf{MLPC per Label Count Is Comparable Across Datasets, Though Subtle Differences Remain}}

\begin{figure*}[!thpb]
    \centering
    \begin{minipage}[b]{0.47\textwidth}
        \includegraphics[width=\textwidth]{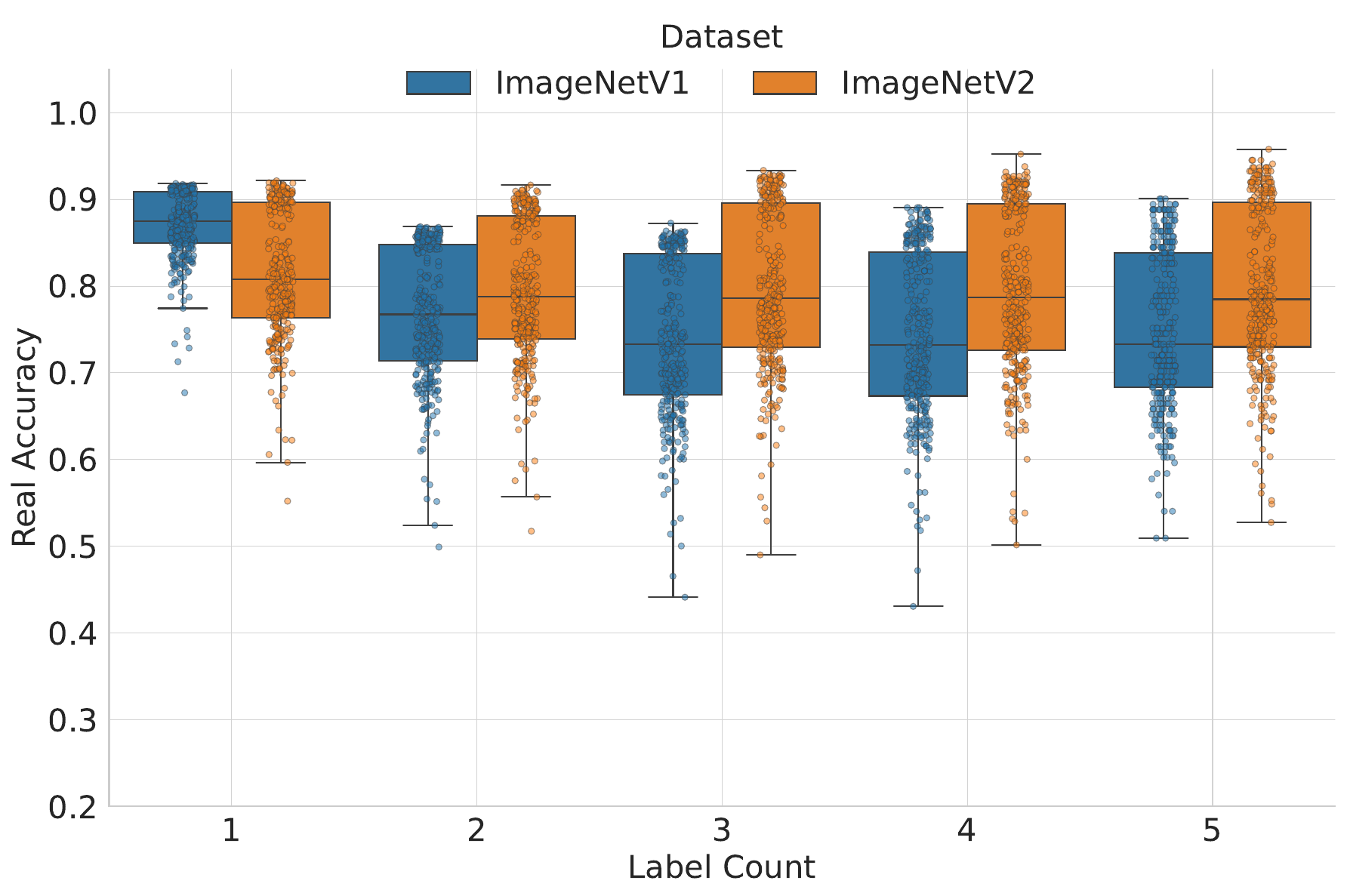}
        \subcaption{}
        \label{fig:real_acc_v1_v2}
    \end{minipage}
    \begin{minipage}[b]{0.47\textwidth}
        \includegraphics[width=\textwidth]{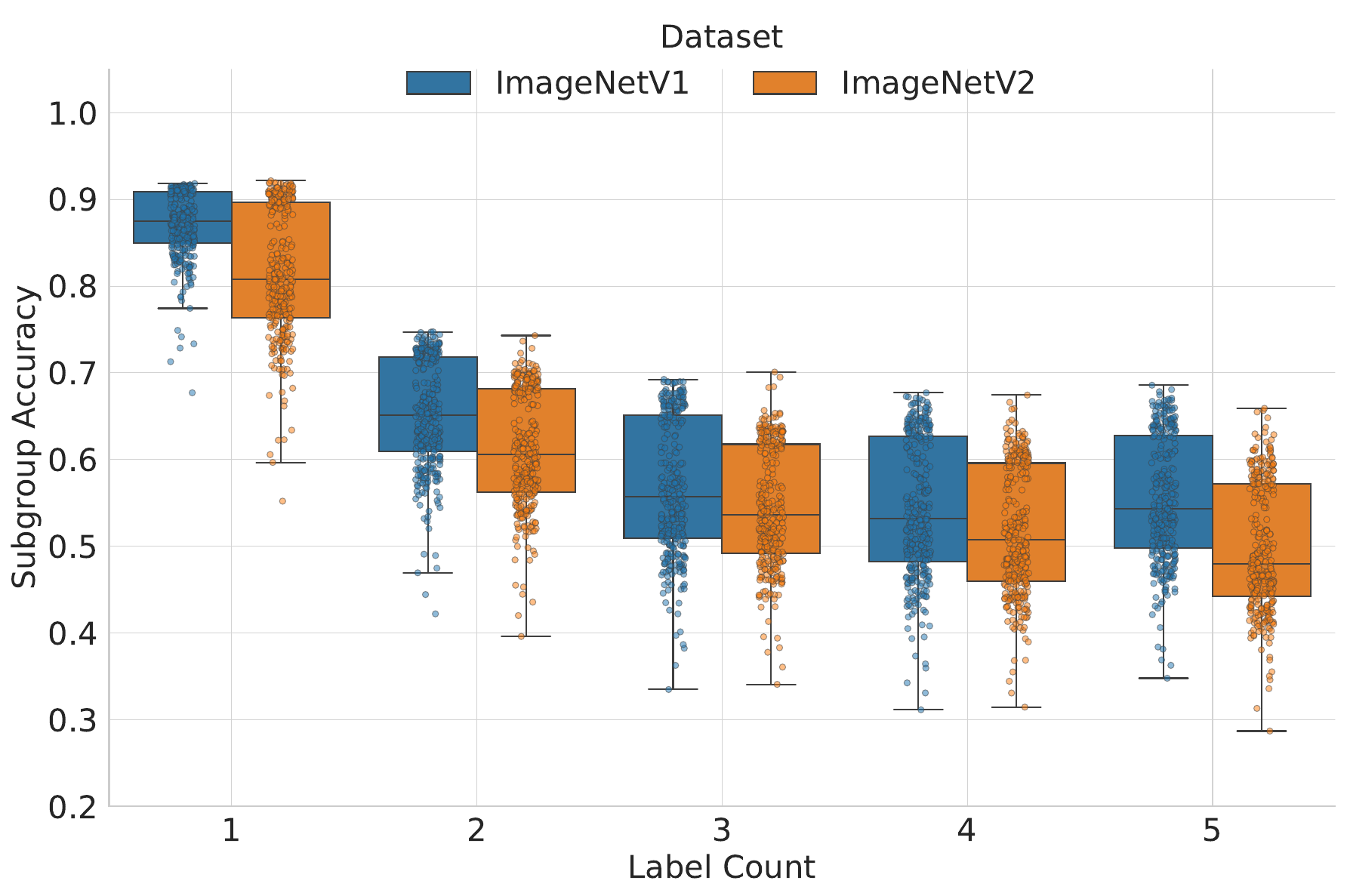}
        \subcaption{}
        \label{fig:subgroup_acc_v1_and_v2}
    \end{minipage}
    \caption{Effectiveness distribution of 315 DNNs pre-trained on ImageNet, with effectiveness obtained on ImageNetV1 and ImageNetV2 (for datapoints with 1-5 labels): (a) ReaL accuracy and (b) Subgroup accuracy versus Label Count. Each dot represents the accuracy of a model for images with the corresponding label count.}
    \label{fig:real_subgroup_acc_on_V1_and_V2}

\end{figure*}

ReaL accuracy and ASMA provide aggregate views of model effectiveness under multi-label evaluation. To refine this understanding, we analyze MLPC as a function of label count in Figure~\ref{fig:real_subgroup_acc_on_V1_and_V2}. Each boxplot shows the distribution of subgroup accuracies across models for a given label count, plotted separately for ImageNetV1 and ImageNetV2. Each point represents the subgroup accuracy of an individual model evaluated on the subset of images with a specific label count. Each box summarizes the distribution of these accuracies across all models at a fixed label count.

Several important observations follow. First, the distributions of ReaL accuracy per label count are broadly similar across the two datasets, with ImageNetV2 showing a slightly higher median accuracy for images containing two or more labels. Second, although the overall trend in subgroup accuracy mirrors that of ReaL accuracy, ImageNetV2 exhibits a lower median accuracy in most cases, indicating that the accuracy drop is not entirely explained by differences in the proportion of multi-label images. Third, MLPC varies considerably across models, as shown by the vertical spread of subgroup accuracies in Figure~\ref{fig:real_subgroup_acc_on_V1_and_V2}.

Overall, the apparent accuracy gap between ImageNetV1 and ImageNetV2 is substantially reduced when evaluation metrics better reflect the multi-label structure. Nevertheless, a difference remains even after accounting for label count, as we briefly discuss in Section~\ref{sec:limitations}.
Finally, Table~\ref{table:label_count_imagenet_dsets} shows that images with more than two labels are relatively infrequent, resulting in an imbalanced distribution across label counts. This imbalance may increase evaluation uncertainty. To account for this and further assess MLPC under controlled label configurations, we extend our analysis using the PatchML dataset in Section~\ref{subsubsec:patchml_mlpc}.


\subsubsection{MLPC of Pre-Trained Models on the PatchML}
\label{subsubsec:patchml_mlpc}

\begin{figure}[!htbp]
    \centering
    \includegraphics[width=0.75\linewidth]{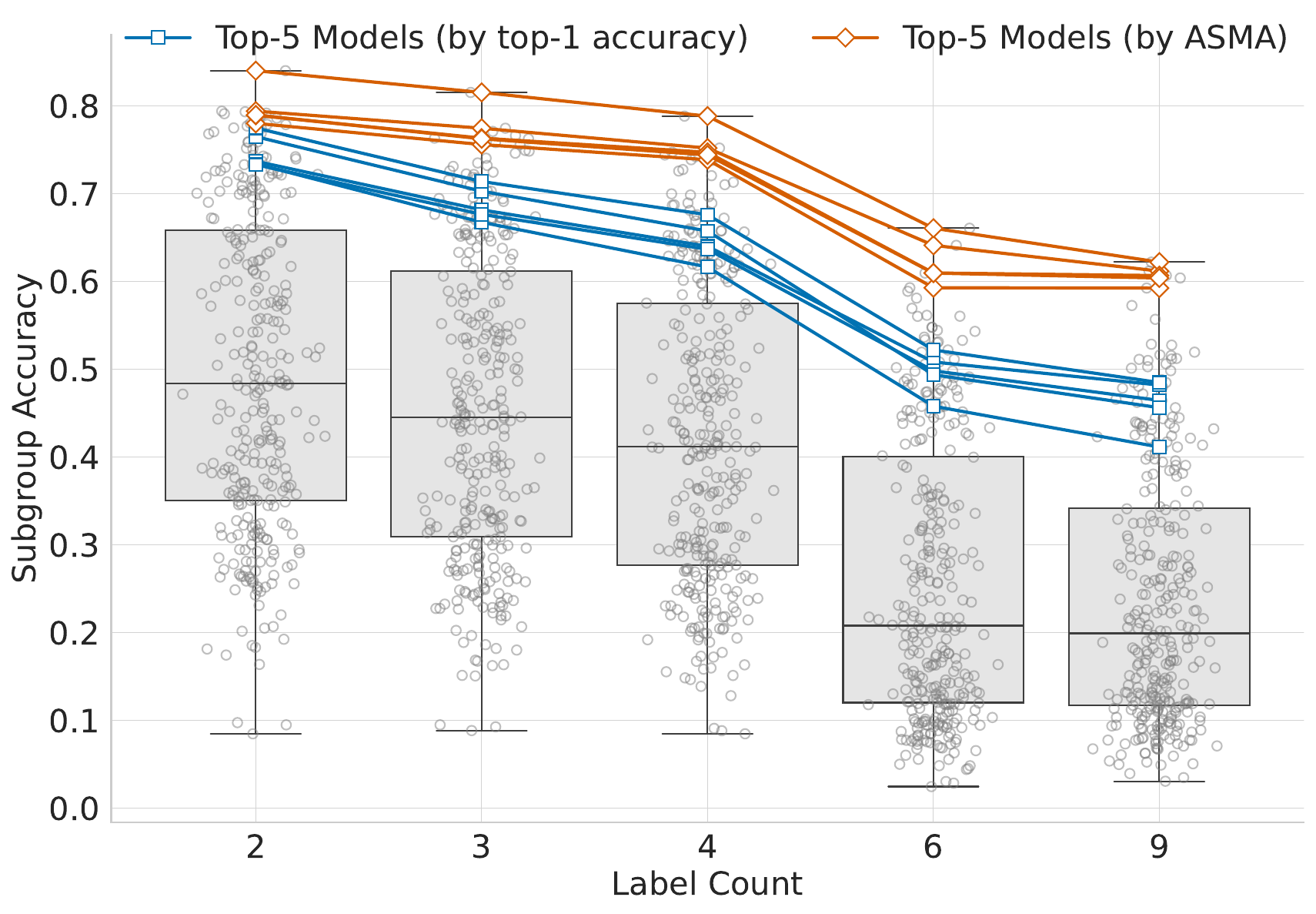}
    \caption{Subgroup accuracy boxplot for $315$ pre-trained DNNs evaluated on PatchML. Overlaid are line plots for the five top-performing models based on ImageNetV1's top-1 accuracy (blue) and PatchML's ASMA (orange).}
    \label{fig:patchml_mlpc}
\end{figure}

\paragraph{\textbf{The PatchML Dataset Confirms the MLPC of ImageNet Pre-trained DNNs}}
We present a boxplot of subgroup accuracy versus label count in Figure~\ref{fig:patchml_mlpc}. Each dot represents the average subgroup accuracy of a model across the five PatchML datasets, each generated using a different seed. This results in $315$ dots per label count, corresponding to the number of evaluated models. The plot demonstrates that models pre-trained on single-label tasks can reasonably predict multiple labels, highlighting their MLPC. Furthermore, this figure confirms the results shown in Figure~\ref{fig:real_subgroup_acc_on_V1_and_V2}, which illustrate the wide range of MLPC between models. Compared to Figure~\ref{fig:real_subgroup_acc_on_V1_and_V2}, Figure~\ref{fig:patchml_mlpc} shows lower median accuracies, potentially due to differences in label distribution and the disruption of semantic relationships between labels in the PatchML datasets.

\paragraph{\textbf{Top-1 Accuracy Masks DNNs with Desirable MLPC Properties}}
Standard benchmarking with top-$1$ accuracy prioritizes single-label correctness, overlooking an important capability: the ability of a model to predict multiple valid labels per image. This limitation is particularly relevant as real-world images often contain multiple objects or concepts, making multi-label recognition a valuable trait for both research and deployment.

We present two key comparisons to illustrate this issue. First, Figure~\ref{fig:patchml_mlpc} overlays line plots on the boxplots, comparing rankings of the same $315$ models based on ImageNetV1 top-1 accuracy and PatchML ASMA. 
The divergence between these rankings highlights a simple point: the top models by ImageNetV1 top-1 accuracy are not the top models by ASMA, as measured on the diagnostic PatchML dataset.
Second, Table~\ref{tab:top_asma_checkpoints} ranks the top-10 models based on PatchML ASMA, listing their top-1 accuracy, ASMA scores, top-1 ranks, and the shift in ranking between top-1 and ASMA. The table shows that models with strong MLPC (high ASMA scores) do not necessarily rank among the top-performing models under top-1 accuracy, reinforcing the need for multi-label-aware evaluation to capture broader model effectiveness. The ASMA presented in the table is the average ASMA for each model across the five PatchML datasets assessed.

For researchers, this finding suggests that advancing DNNs purely through top-1 accuracy may neglect improvements in multi-label recognition, which could be crucial for applications requiring broader scene understanding. For practitioners, these results emphasize that selecting models based solely on top-1 accuracy may lead to suboptimal choices for real-world tasks where multiple relevant categories co-occur. Incorporating MLPC evaluation provides a more complete picture of a model’s image recognition capabilities, ensuring that models are evaluated based on criteria aligned with practical deployment scenarios.

\begin{table}[htbp]
\centering
\footnotesize
\caption{DNN effectiveness and rank changes of the top-$10$ pre-trained models based on ASMA on PatchML, listed in descending order of ASMA (best at top). $\Delta$Rank indicates the ranking shift between top-1 accuracy on ImageNetV1 and ASMA on PatchML.}

\label{tab:top_asma_checkpoints}
\begin{tabular}{l
                @{\hspace{0.5em}}
                c
                @{\hspace{1em}}
                c c
                @{\hspace{1em}}
                c}
\toprule
\multirow{2}{*}{\textbf{Pre-Trained Models}} &
\multicolumn{1}{c}{\textbf{PatchML}} &
\multicolumn{2}{c}{\textbf{ImageNetV1}} &
\multirow{2}{*}{\textbf{$\Delta$ Rank}} \\
\cmidrule(lr){2-2} \cmidrule(lr){3-4}
& \textbf{ASMA} & \textbf{Top-$1$} & \textbf{Top-$1$ Rank} & \\
\midrule
eva\_large\_patch14\_336.in22k\_ft\_in1k  & 74.50 & 88.50 & 14  & +13 \\
convnextv2\_huge.fcmae\_ft\_in22k\_in1k\_512  & 71.45 & 88.84 & 8  & +6 \\
volo\_d5\_448\_in1k  & 70.30 & 87.05 & 56  & +53  \\
volo\_d5\_512\_in1k  & 70.17 & 87.04 & 57  & +53  \\
volo\_d4\_448\_in1k  & 69.18 & 86.86 & 63  & +58 \\
volo\_d3\_448\_in1k  & 69.16 & 86.60 & 69  & +63 \\
convnextv2\_huge.fcmae\_ft\_in22k\_in1k\_384  & 67.92 & 88.60 & 10  & +3 \\
eva\_large\_patch14\_196.in22k\_ft\_in1k  & 67.20 & 87.77 & 36  & +28 \\
beitv2\_large\_patch16\_224.in1k\_ft\_in1k  & 67.16 & 87.23 & 51 & +42 \\
cait\_m48\_448.fb\_dist\_in1k  & 67.11 & 86.32 & 81  & +71 \\
\bottomrule
\end{tabular}
\end{table}

\paragraph{\textbf{Top ASMA Models Reveal Architectural and Training Factors Underpinning Multi-Label Generalization}}

Although the primary goal of this study is not to rank models by their MLPC, we identify and analyze the ten top-performing models based on ASMA to uncover architectural and training-related factors that may underlie their multi-label robustness. All ten models exhibit less than a $1\%$ performance drop under ReaL accuracy, but only four models fall below this threshold under ASMA, and all four are VOLO variants~\citep{yuan_volo_2023}. Interestingly, VOLO-based models are the only ones in this group that our research confirms do not leverage external datasets beyond ImageNet-1K. They are also the only models to simultaneously close the gap below $1\%$ under both ReaL and ASMA, highlighting their distinctive training dynamics.

VOLO is a family of ``Vision Outlooker'' architectures that improve spatial modeling in vision transformers using a lightweight outlooker module. While trained solely on ImageNet-1K, VOLO inherits the token labeling framework from LV-ViT~\citep{jiang_all_2021}, where patch-wise supervision is derived from a dense score map generated by a high-capacity convolutional model, specifically NFNet-F6, which itself is trained only on ImageNet-1K. This dense labeling provides spatially aligned pseudo-labels to each patch token, enabling the model to receive fine-grained semantic supervision beyond the standard single-label target. Although LV-ViT does not explicitly frame token labeling as knowledge distillation, the transfer of structured semantic maps from a pretrained teacher resembles offline distillation in form and effect. As a result, VOLO models are implicitly exposed to richer intra-image semantics during training, which may contribute to their ability to generalize under label-incomplete test conditions.

ConvNeXtV2~\citep{woo_convnext_2023}, the only fully convolutional model in the top ten, follows a different path. It builds upon ConvNeXt~\citep{liu_convnet_2022} and introduces Global Response Normalization to improve feature selectivity and suppress channel collapse. Pre-trained using FCMAE on full ImageNet with approximately $22,000$ classes, ConvNeXtV2 learns to reconstruct masked image regions without relying on attention. This self-supervised pretraining encourages spatially complete and semantically robust representations, aiding its ability to recognize multiple co-occurring objects even under traditional single-label fine-tuning.

Several top ASMA models, including EVA~\citep{sun_eva-clip_2023} and BEiT v2~\citep{peng_beit_2022}, leverage MIM with supervision derived from CLIP~\citep{radford_learning_2021}. BEiT v2 constructs its MIM targets using discrete visual tokens obtained from a CLIP-based vision-language model, requiring the model to reconstruct these semantic units from masked inputs. EVA, in contrast, discards tokenization and directly regresses high-level CLIP features, integrating the semantic richness of CLIP with the geometric inductive biases of masked prediction. Despite these differences, both models combine MIM objectives with large-scale vision-text corpora; they plausibly gain multi-label capabilities by internalizing object co-occurrence patterns and context from broader visual-textual associations.

CaiT-M48~\citep{touvron_going_2021} provides yet another route to strong multi-label performance by combining extremely deep transformer stacks, class-attention modules, and knowledge distillation. The model is distilled from a RegNetY-16GF teacher trained via Facebook’s SWAG pipeline~\citep{radosavovic_designing_2020}, a semi-weakly supervised framework that first pre-trains on 3.6 billion public Instagram images using noisy hashtag labels, followed by fine-tuning on ImageNet-1K. This two-stage process enables the teacher to distill broad semantic knowledge from web-scale data, while the student is trained exclusively on ImageNet-1K. During distillation, this knowledge, potentially encoding latent co-occurrence statistics and rich contextual patterns, is transferred to CaiT-M48. Its architecture, particularly the class-attention layers, is well-suited to refine these features for downstream classification. This form of indirect supervision, mediated through a high-capacity teacher with web-enhanced pretraining, illustrates how multi-label-relevant semantics could be inherited even when the student model is trained strictly within the ImageNet-1K protocol.

Together, these top ASMA models exhibit a unifying pattern: they are all trained with supervision signals that transcend the single-label-per-image constraint. Whether via dense token-level pseudo-labels (VOLO), semantic visual tokens (BEiT v2), high-level vision-language features (EVA), masked reconstruction (EVA, BEiT v2, ConvNeXtV2), or deep teacher knowledge (CaiT), each model leverages training signals that align more closely with the inherently multi-label nature of real-world images. These signals encourage the learning of compositional and object-aware features that traditional top-1 accuracy fails to capture; ReaL accuracy only partially reflects them.

Furthermore, these models sustain their multi-label generalization across both ImageNetV1 and ImageNetV2, suggesting that their semantic competence is not an artifact of dataset-specific biases. ASMA captures this cross-distribution consistency more effectively than traditional metrics, reinforcing that genuine multi-label robustness arises not from scale alone but from embedding richer label structure during early representation learning.

In conclusion, our preliminary analysis suggests that high MLPC is not simply a function of model size or resolution, but is driven by training strategies that incorporate latent or auxiliary supervision. These strategies enable models to move beyond single-label assumptions and recover fuller object semantics, often without ever being explicitly trained to do so. This calls into question the heavy reliance on top-$1$ accuracy benchmarks and highlights the need for multi-label-aware evaluation in assessing semantic completeness.

\paragraph{\textbf{Example Predictions for High Label Counts Confirm MLPC in DNNs}}
The preceding results demonstrate that single-label pre-trained ImageNet models exhibit MLPC, with subgroup accuracy analysis (Figure~\ref{fig:patchml_mlpc}) indicating that models successfully recognize multiple objects per image to varying extents.   
To further validate this capability, we examine predictions on images containing nine distinct objects---the highest label count simulated in PatchML.
 This setting presents a stringent test of MLPC, requiring models to accurately prioritize multiple valid labels while minimizing misclassification.

\begin{figure*}[!htbp]
    \centering

    \begin{subfigure}[b]{0.48\textwidth}
        \centering
        \includegraphics[width=\linewidth]{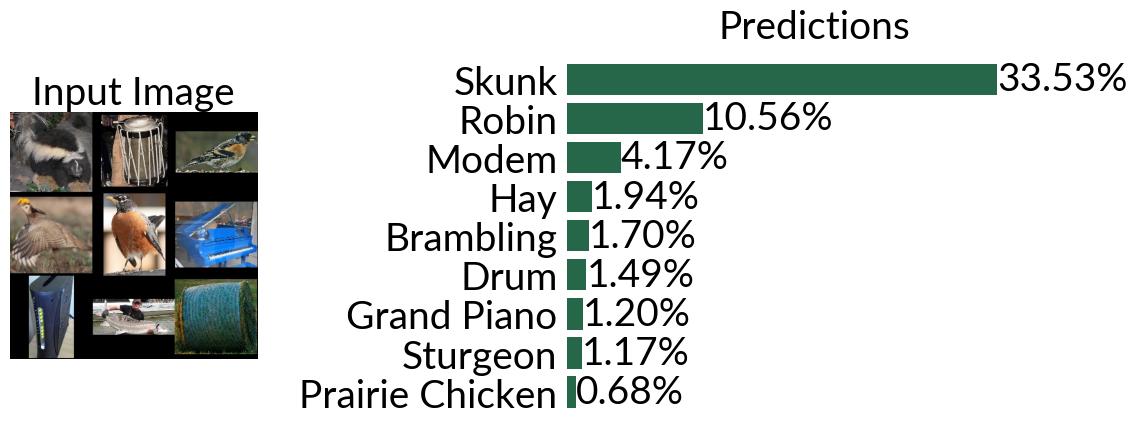}
        \caption{All top-$9$ predictions exactly match the ground labels.}
        \label{fig:labela}
    \end{subfigure}
    \hfill
    \begin{subfigure}[b]{0.48\textwidth}
        \centering
        \includegraphics[width=\linewidth]{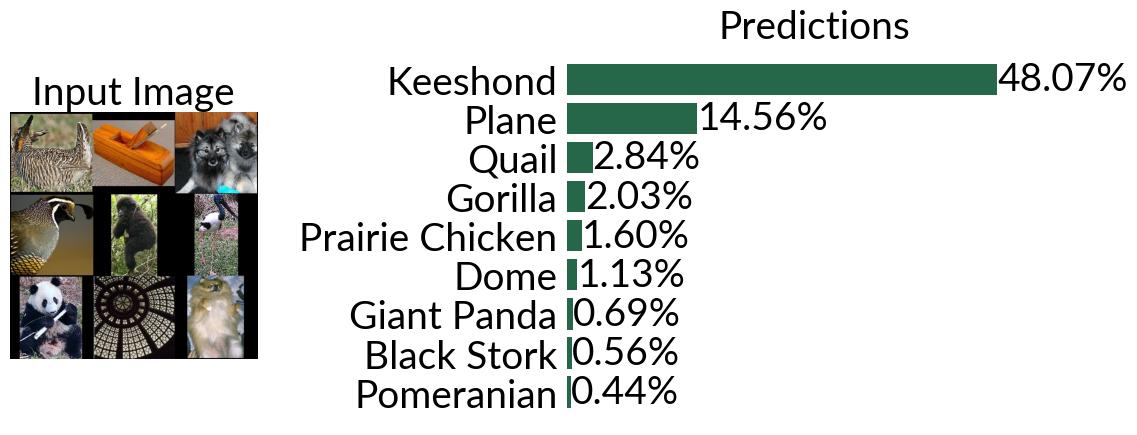}
        \caption{All top-$9$ predictions exactly match the ground labels.}
        \label{fig:labelb}
    \end{subfigure}

    \vspace{0.5cm} 

    \begin{subfigure}[b]{0.48\textwidth}
        \centering
        \includegraphics[width=\linewidth]{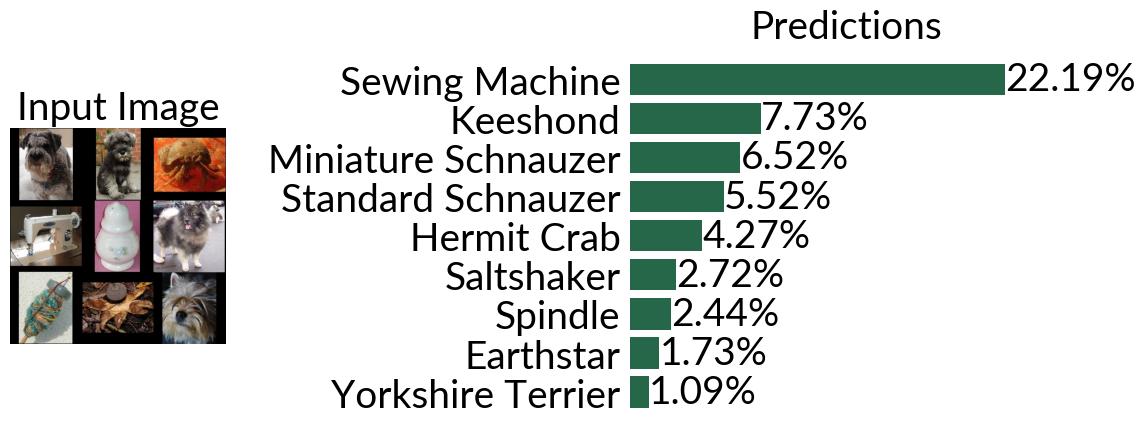}
        \caption{All top-$9$ predictions exactly match the ground labels.}
        \label{fig:labelc}
    \end{subfigure}
    \hfill
    \begin{subfigure}[b]{0.48\textwidth}
        \centering
        \includegraphics[width=\linewidth]{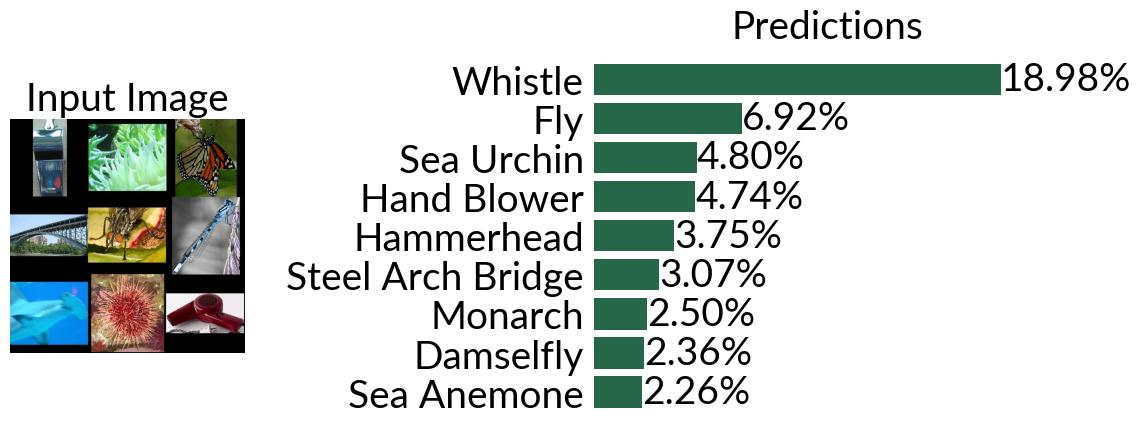}
        \caption{All top-$9$ predictions exactly match the ground labels.}
        \label{fig:labeld}
    \end{subfigure}

    \vspace{0.5cm} 

    \begin{subfigure}[b]{0.48\textwidth}
        \centering
        \includegraphics[width=\linewidth]{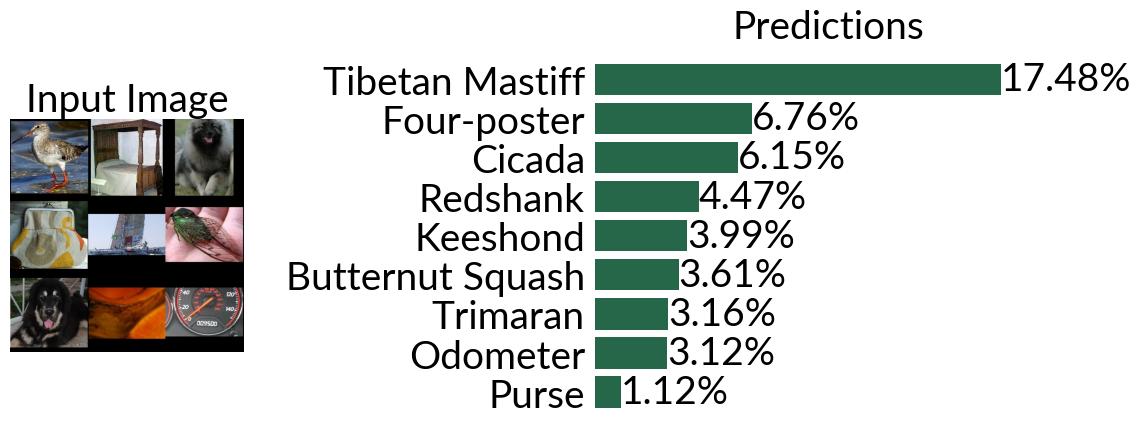}
        \caption{All top-$9$ predictions exactly match the ground labels.}
        \label{fig:label5}
    \end{subfigure}
    \hfill
    \begin{subfigure}[b]{0.48\textwidth}
        \centering
        \includegraphics[width=\linewidth]{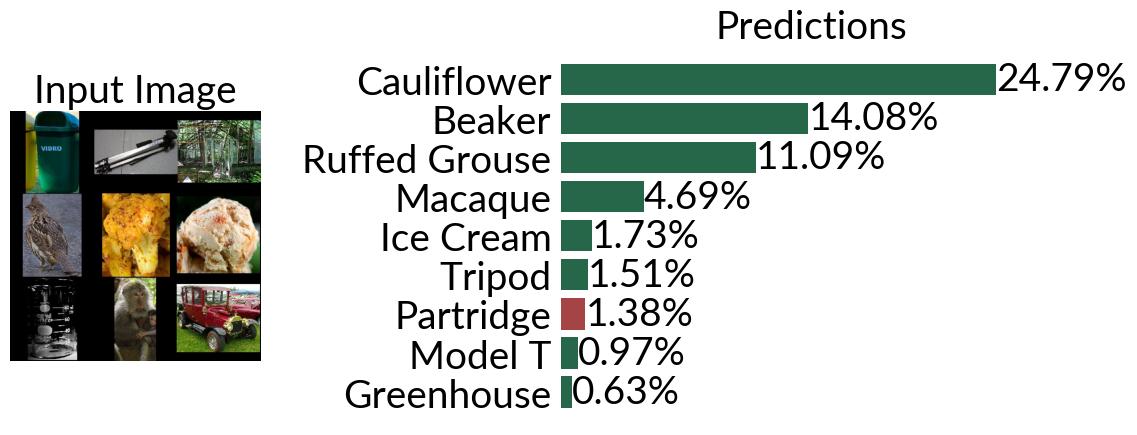}
        \caption{The ground truth label not in the top-$9$ is \emph{Ash Can}.}
        \label{fig:label6}
    \end{subfigure}

    \vspace{0.5cm} 

    \begin{subfigure}[b]{0.48\textwidth}
        \centering
        \includegraphics[width=\linewidth]{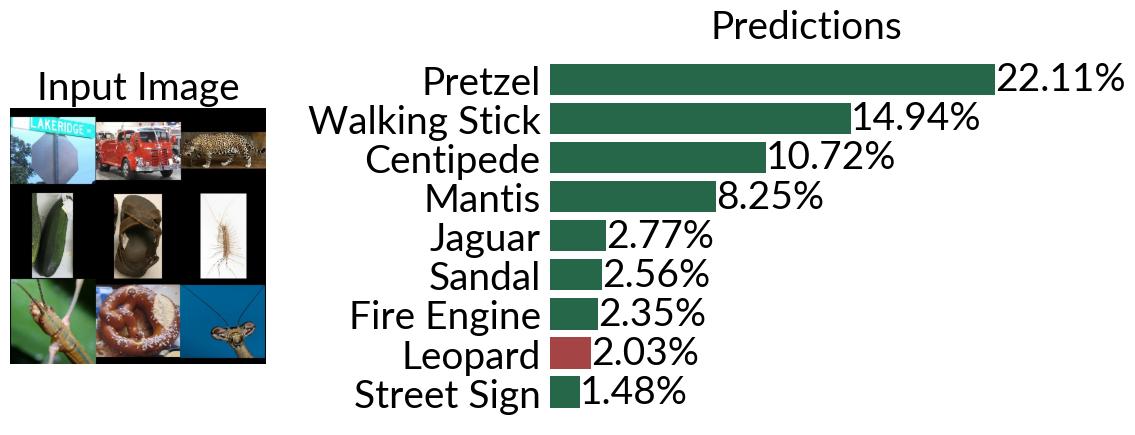}
        \caption{The ground truth label not in the top-$9$ is \emph{Zucchini}.}
        \label{fig:label7}
    \end{subfigure}
    \hfill
    \begin{subfigure}[b]{0.48\textwidth}
        \centering
        \includegraphics[width=\linewidth]{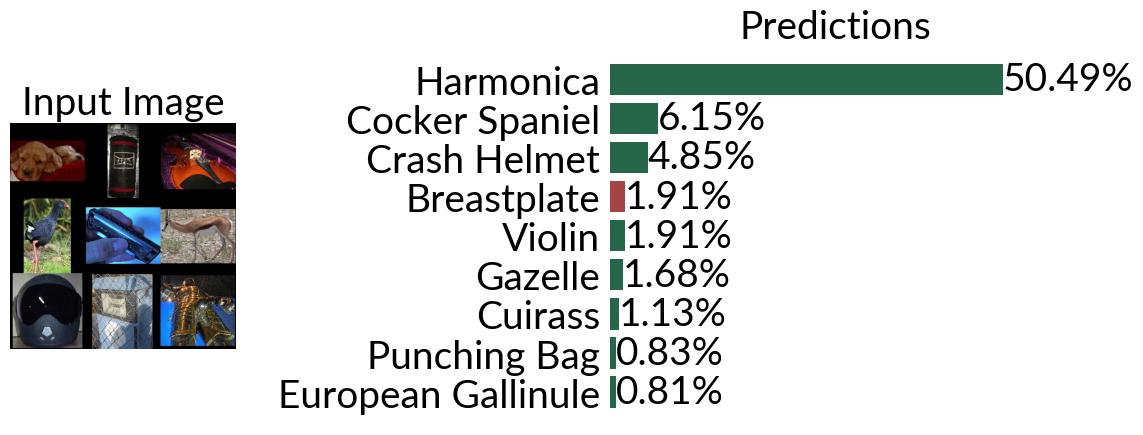}
        \caption{The ground truth label not in the top-$9$ is \emph{Gas Pump}.}
        \label{fig:label8}
    \end{subfigure}

    \vspace{0.5cm} 

    \begin{subfigure}[b]{0.48\textwidth}
        \centering
        \includegraphics[width=\linewidth]{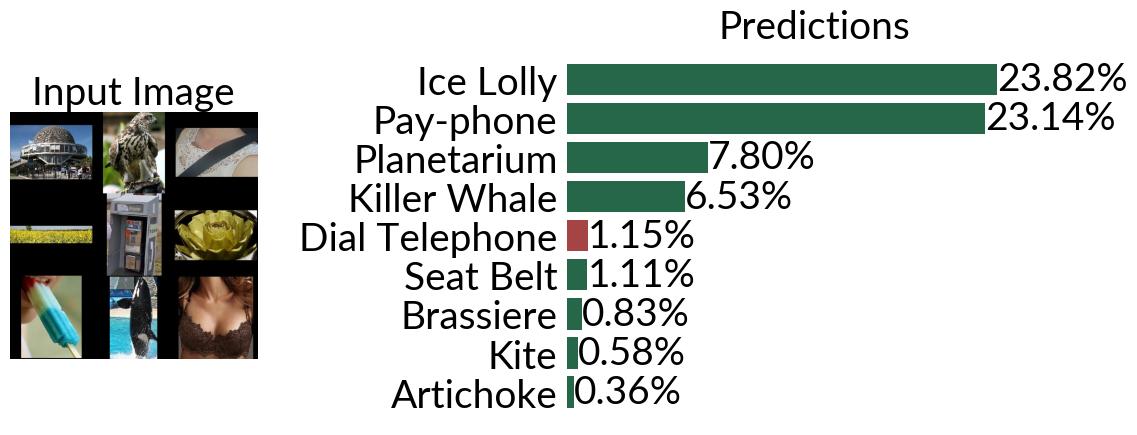}
        \caption{The ground truth label not in the top-$9$ is \emph{Rapeseed}.}
        \label{fig:label9}
    \end{subfigure}
    \hfill
    \begin{subfigure}[b]{0.48\textwidth}
        \centering
        \includegraphics[width=\linewidth]{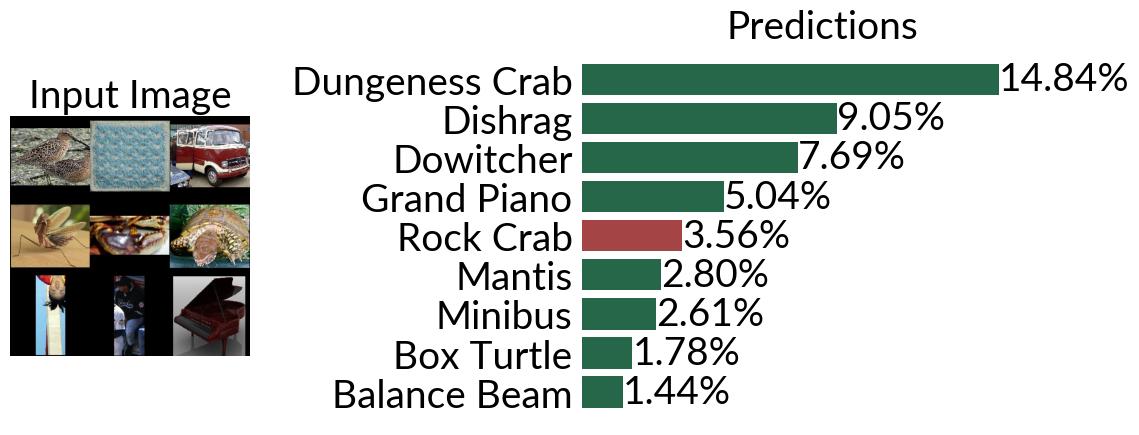}
        \caption{The ground truth label not in the top-$9$ is \emph{Ballplayer}.}
        \label{fig:label10}
    \end{subfigure}

    \caption{Examples of PatchML predictions for sample images with nine ground truth labels. Red bars indicate predicted labels that are not part of the ground truth set.}
    \label{fig:patchml_label_count_9_examples}
\end{figure*}

Figure~\ref{fig:patchml_label_count_9_examples} presents representative predictions from the top-performing model identified in Table~\ref{tab:top_asma_checkpoints}, \texttt{eva\_large\_patch14\_336.in22k\_ft\_in1k}. Several observations emerge: ($i$) the model correctly ranks most valid labels while distributing softmax probabilities across multiple objects, and ($ii$) the model does not exhibit a clear case of center bias, a tendency reported in prior work~\citep{kruthiventi_deepfix_2015, jetley_end--end_2016, taesiri_imagenet-hard_2023}, where a model favors objects near the image center due to dataset artifacts or augmentation-induced biases (that is, the model does not consistently assign higher probabilities to the objects in the middle). These findings further reinforce that, despite being trained under a single-label framework, ImageNet models demonstrate substantial MLPC, highlighting the need for evaluation metrics that explicitly account for this capability.

\section{Limitations and Future Directions} 
\label{sec:limitations}

This study establishes a principled evaluation framework for assessing the MLPC of single-label-trained ImageNet models. By leveraging refined annotations, variable top-$k$ evaluation, the ASMA metric, and PatchML, we demonstrate that many DNNs possess latent multi-label capabilities that are systematically underrepresented by standard evaluation protocols. Our large-scale analysis, conducted on a broad pool of models, provides new insights into the causes of the ImageNetV2 accuracy degradation and highlights how more nuanced benchmarking can reveal strengths in model behavior that are otherwise obscured. These contributions offer a robust foundation for reevaluating model generalization and reliability in vision benchmarks.

Despite these strengths, several limitations define the scope of our findings and present opportunities for future research.

First, our analysis relies on the availability of existing re-annotated multi-label ground truths for ImageNetV1~\citep{beyer_are_2020} and ImageNetV2~\citep{anzaku_leveraging_2024}. While these annotations offer substantially improved label coverage, they reflect the specific protocols used in their construction and may still miss plausible object categories, particularly in cases involving fine-grained distinctions, occlusion, or ambiguous context. Future work could explore consensus-driven or ontology-aware annotation pipelines to further enrich evaluation datasets and quantify residual uncertainty in the label space.

Second, the variable top-$k$ evaluation approach assumes that a model capable of recognizing all valid objects in an image will rank them among its top-$k$ predictions. While this is a reasonable assumption for retrospective analysis of MLPC, it is not intended as a method for converting single-label models into real-world multi-label classifiers. Future research may investigate how training paradigms could be adapted to bridge this gap, such as by studying how top-$k$ prediction behavior evolves under different loss functions or supervision regimes.

Third, while the PatchML dataset serves as a valuable diagnostic tool by removing contextual and co-occurrence cues, its synthetic nature imposes constraints on generalizability. Natural scenes often include semantically meaningful spatial relationships, textures, and occlusions that interact with object recognition. Extending PatchML to incorporate controlled context variation could provide a richer testbed for studying model robustness to both object combinations and scene-level semantics.

Finally, this work strictly focuses on evaluation. All models were trained using standard single-label objectives, and our results reveal their latent MLPC under these constraints. Although we conducted a preliminary analysis of the top models' architectural and training characteristics, a deeper investigation into the causal links between model design and multi-label behavior remains unexplored. Furthermore, future work could assess whether ASMA correlates with performance on downstream tasks such as object detection, segmentation, or out-of-distribution detection. Understanding these relationships would help determine whether MLPC serves as a transferable property relevant to trustworthy model selection and deployment.

Together, these limitations identify fruitful directions for extending the present framework while reaffirming the broader value of this study: single-label evaluation metrics alone are no longer sufficient for diagnosing the full spectrum of model behavior. Addressing these gaps will be critical as the field moves toward more holistic assessments of reliability in vision systems.


\section{Conclusions}

ImageNet remains a fundamental benchmark for computer vision research, particularly for evaluating model capabilities beyond single-label classification.
Ensuring that ImageNet benchmarking accurately reflects the nuanced capabilities of DNNs is crucial to steer research in the right direction. In this work, we critically examine the prevailing single-label assumption in ImageNet benchmarking and its role in the reported unexpected and largely unexplained degradation in effectiveness observed in ImageNetV2. By embracing a multi-label perspective, we found that a critical understudied factor in the reported accuracy degradation is the difference in the proportion of multi-label datapoints between ImageNetV1 and ImageNetV2, a finding that has been obscured by the typical use of a single-label evaluation approach. Our experimental results did not find support for the reported substantial degradation in effectiveness observed in ImageNetV2. In contrast, we found that ImageNet pre-trained DNNs, on average, perform equally well or even better on ImageNetV2. Furthermore, our analysis reveals that single-label evaluation fails to capture the inherent complexity of the ImageNet dataset. By solely focusing on single-label classification, there is a risk of drawing incomplete or even misleading conclusions about the effectiveness of DNNs. Such oversights may obscure a comprehensive understanding of DNN capabilities and could inadvertently redirect research efforts away from crucial considerations surrounding the reliability and generalizability of DNNs. Therefore, we advocate for the complementary adoption of a multi-label evaluation approach in future ImageNet benchmarking practices, believing that this approach will foster advancements in computer vision models that are more aligned with the complexities of the real world.

\section{Acknowledgments}
I confirm that this research did not receive any specific grant from funding agencies in the public, commercial, or not-for-profit sectors. The work was conducted as part of my appointment as AAP at Ghent University.

\bibliographystyle{elsarticle-num-names}
\bibliography{references}

\begin{thebibliography}{46}
\expandafter\ifx\csname natexlab\endcsname\relax\def\natexlab#1{#1}\fi
\providecommand{\url}[1]{\texttt{#1}}
\providecommand{\href}[2]{#2}
\providecommand{\path}[1]{#1}
\providecommand{\DOIprefix}{doi:}
\providecommand{\ArXivprefix}{arXiv:}
\providecommand{\URLprefix}{URL: }
\providecommand{\Pubmedprefix}{pmid:}
\providecommand{\doi}[1]{\href{http://dx.doi.org/#1}{\path{#1}}}
\providecommand{\Pubmed}[1]{\href{pmid:#1}{\path{#1}}}
\providecommand{\bibinfo}[2]{#2}
\ifx\xfnm\relax \def\xfnm[#1]{\unskip,\space#1}\fi
\bibitem[{Deng et~al.(2009)Deng, Dong, Socher, Li, Li, and Fei-Fei}]{deng_imagenet_2009}
\bibinfo{author}{J.~Deng}, \bibinfo{author}{W.~Dong}, \bibinfo{author}{R.~Socher}, \bibinfo{author}{L.-J. Li}, \bibinfo{author}{K.~Li}, \bibinfo{author}{L.~Fei-Fei},
\newblock \bibinfo{title}{Imagenet: {A} {Large}-scale {Hierarchical} {Image} {Database}},
\newblock in: \bibinfo{booktitle}{{IEEE} conference on computer vision and pattern recognition}, \bibinfo{year}{2009}, pp. \bibinfo{pages}{248--255}.
\bibitem[{Russakovsky et~al.(2015)Russakovsky, Deng, Su, Krause, Satheesh, Ma, Huang, Karpathy, Khosla, Bernstein, Berg, and Fei-Fei}]{russakovsky_imagenet_2015}
\bibinfo{author}{O.~Russakovsky}, \bibinfo{author}{J.~Deng}, \bibinfo{author}{H.~Su}, \bibinfo{author}{J.~Krause}, \bibinfo{author}{S.~Satheesh}, \bibinfo{author}{S.~Ma}, \bibinfo{author}{Z.~Huang}, \bibinfo{author}{A.~Karpathy}, \bibinfo{author}{A.~Khosla}, \bibinfo{author}{M.~Bernstein}, \bibinfo{author}{A.~C. Berg}, \bibinfo{author}{L.~Fei-Fei},
\newblock \bibinfo{title}{{ImageNet} {Large} {Scale} {Visual} {Recognition} {Challenge}},
\newblock \bibinfo{journal}{International Journal of Computer Vision} \bibinfo{volume}{115} (\bibinfo{year}{2015}) \bibinfo{pages}{211--252}. \DOIprefix\doi{10.1007/s11263-015-0816-y}.
\bibitem[{Barbu et~al.(2019)Barbu, Mayo, Alverio, Luo, Wang, Gutfreund, Tenenbaum, and Katz}]{barbu_objectnet_2019}
\bibinfo{author}{A.~Barbu}, \bibinfo{author}{D.~Mayo}, \bibinfo{author}{J.~Alverio}, \bibinfo{author}{W.~Luo}, \bibinfo{author}{C.~Wang}, \bibinfo{author}{D.~Gutfreund}, \bibinfo{author}{J.~Tenenbaum}, \bibinfo{author}{B.~Katz},
\newblock \bibinfo{title}{{ObjectNet}: {A} large-scale bias-controlled dataset for pushing the limits of object recognition models},
\newblock in: \bibinfo{booktitle}{Advances in {Neural} {Information} {Processing} {Systems}}, volume~\bibinfo{volume}{32}, \bibinfo{year}{2019}.
\bibitem[{Recht et~al.(2019)Recht, Roelofs, Schmidt, and Shankar}]{recht_imagenet_2019}
\bibinfo{author}{B.~Recht}, \bibinfo{author}{R.~Roelofs}, \bibinfo{author}{L.~Schmidt}, \bibinfo{author}{V.~Shankar},
\newblock \bibinfo{title}{Do {ImageNet} {Classifiers} {Generalize} to {ImageNet}?},
\newblock in: \bibinfo{editor}{K.~Chaudhuri}, \bibinfo{editor}{R.~Salakhutdinov} (Eds.), \bibinfo{booktitle}{Proceedings of the 36th {International} {Conference} on {Machine} {Learning}}, volume~\bibinfo{volume}{97}, \bibinfo{year}{2019}, pp. \bibinfo{pages}{5389--5400}.
\bibitem[{Beyer et~al.(2020)Beyer, Hénaff, Kolesnikov, Zhai, and Oord}]{beyer_are_2020}
\bibinfo{author}{L.~Beyer}, \bibinfo{author}{O.~J. Hénaff}, \bibinfo{author}{A.~Kolesnikov}, \bibinfo{author}{X.~Zhai}, \bibinfo{author}{A.~v.~d. Oord}, \bibinfo{title}{Are we done with {ImageNet}?}, \bibinfo{year}{2020}. \URLprefix \url{http://arxiv.org/abs/2006.07159}.
\bibitem[{Shankar et~al.(2020)Shankar, Roelofs, Mania, Fang, Recht, and Schmidt}]{shankar_evaluating_2020}
\bibinfo{author}{V.~Shankar}, \bibinfo{author}{R.~Roelofs}, \bibinfo{author}{H.~Mania}, \bibinfo{author}{A.~Fang}, \bibinfo{author}{B.~Recht}, \bibinfo{author}{L.~Schmidt},
\newblock \bibinfo{title}{Evaluating {Machine} {Accuracy} on {ImageNet}},
\newblock in: \bibinfo{booktitle}{International {Conference} on {Machine} {Learning}}, volume~\bibinfo{volume}{37}, \bibinfo{year}{2020}, pp. \bibinfo{pages}{8634--8644}.
\bibitem[{Tsipras et~al.(2020)Tsipras, Santurkar, Engstrom, Ilyas, and Madry}]{tsipras_imagenet_2020}
\bibinfo{author}{D.~Tsipras}, \bibinfo{author}{S.~Santurkar}, \bibinfo{author}{L.~Engstrom}, \bibinfo{author}{A.~Ilyas}, \bibinfo{author}{A.~Madry},
\newblock \bibinfo{title}{From {ImageNet} to {Image} {Classification}: {Contextualizing} {Progress} on {Benchmarks}},
\newblock in: \bibinfo{booktitle}{International {Conference} on {Machine} {Learning}}, \bibinfo{year}{2020}, pp. \bibinfo{pages}{9625--9635}.
\bibitem[{Vasudevan et~al.(2022)Vasudevan, Caine, Gontijo-Lopes, Fridovich-Keil, and Roelofs}]{vasudevan_when_2022}
\bibinfo{author}{V.~Vasudevan}, \bibinfo{author}{B.~Caine}, \bibinfo{author}{R.~Gontijo-Lopes}, \bibinfo{author}{S.~Fridovich-Keil}, \bibinfo{author}{R.~Roelofs},
\newblock \bibinfo{title}{When does dough become a bagel? {Analyzing} the remaining mistakes on {ImageNet}},
\newblock in: \bibinfo{booktitle}{Conference on {Neural} {Information} {Processing} {Systems}}, \bibinfo{year}{2022}.
\bibitem[{Kisel et~al.(2024)Kisel, Volkov, Hanzelkova, Janouskova, and Matas}]{kisel_flaws_2024}
\bibinfo{author}{N.~Kisel}, \bibinfo{author}{I.~Volkov}, \bibinfo{author}{K.~Hanzelkova}, \bibinfo{author}{K.~Janouskova}, \bibinfo{author}{J.~Matas}, \bibinfo{title}{Flaws of {ImageNet}, {Computer} {Vision}'s {Favourite} {Dataset}}, \bibinfo{year}{2024}. \DOIprefix\doi{10.48550/arXiv.2412.00076}.
\bibitem[{Northcutt et~al.(2021)Northcutt, Athalye, and Mueller}]{northcutt_pervasive_2021}
\bibinfo{author}{C.~G. Northcutt}, \bibinfo{author}{A.~Athalye}, \bibinfo{author}{J.~Mueller},
\newblock \bibinfo{title}{Pervasive {Label} {Errors} in {Test} {Sets} {Destabilize} {Machine} {Learning} {Benchmarks}},
\newblock in: \bibinfo{booktitle}{Thirty-fifth {Conference} on {Neural} {Information} {Processing} {Systems} {Datasets} and {Benchmarks} {Track}}, \bibinfo{year}{2021}. \URLprefix \url{https://openreview.net/forum?id=XccDXrDNLek}.
\bibitem[{Luccioni and Rolnick(2023)}]{luccioni_bugs_2023}
\bibinfo{author}{A.~S. Luccioni}, \bibinfo{author}{D.~Rolnick},
\newblock \bibinfo{title}{Bugs in the data: how {ImageNet} misrepresents biodiversity},
\newblock in: \bibinfo{booktitle}{Proceedings of the {Thirty}-{Seventh} {AAAI} {Conference} on {Artificial} {Intelligence} and {Thirty}-{Fifth} {Conference} on {Innovative} {Applications} of {Artificial} {Intelligence} and {Thirteenth} {Symposium} on {Educational} {Advances} in {Artificial} {Intelligence}}, volume~\bibinfo{volume}{37}, \bibinfo{year}{2023}, pp. \bibinfo{pages}{14382--14390}. \DOIprefix\doi{10.1609/aaai.v37i12.26682}.
\bibitem[{Idrissi et~al.(2022)Idrissi, Bouchacourt, Balestriero, Evtimov, Hazirbas, Ballas, Vincent, Drozdzal, Lopez-Paz, and Ibrahim}]{idrissi_imagenet-x_2022}
\bibinfo{author}{B.~Y. Idrissi}, \bibinfo{author}{D.~Bouchacourt}, \bibinfo{author}{R.~Balestriero}, \bibinfo{author}{I.~Evtimov}, \bibinfo{author}{C.~Hazirbas}, \bibinfo{author}{N.~Ballas}, \bibinfo{author}{P.~Vincent}, \bibinfo{author}{M.~Drozdzal}, \bibinfo{author}{D.~Lopez-Paz}, \bibinfo{author}{M.~Ibrahim},
\newblock \bibinfo{title}{{ImageNet}-{X}: {Understanding} {Model} {Mistakes} with {Factor} of {Variation} {Annotations}},
\newblock \bibinfo{year}{2022}.
\bibitem[{Peychev et~al.(2023)Peychev, Müller, Fischer, and Vechev}]{peychev_automated_2023}
\bibinfo{author}{M.~Peychev}, \bibinfo{author}{M.~N. Müller}, \bibinfo{author}{M.~Fischer}, \bibinfo{author}{M.~Vechev},
\newblock \bibinfo{title}{Automated {Classification} of {Model} {Errors} on {ImageNet}},
\newblock \bibinfo{journal}{Thirty-seventh Conference on Neural Information Processing Systems}  (\bibinfo{year}{2023}).
\bibitem[{Anzaku et~al.(2024)Anzaku, Hong, Park, Yang, Kim, Won, Herath, Van~Messem, and De~Neve}]{anzaku_leveraging_2024}
\bibinfo{author}{E.~T. Anzaku}, \bibinfo{author}{H.~Hong}, \bibinfo{author}{J.-W. Park}, \bibinfo{author}{W.~Yang}, \bibinfo{author}{K.~Kim}, \bibinfo{author}{J.~Won}, \bibinfo{author}{D.~V.~K. Herath}, \bibinfo{author}{A.~Van~Messem}, \bibinfo{author}{W.~De~Neve},
\newblock \bibinfo{title}{Leveraging {Human}-{Machine} {Interactions} for {Computer} {Vision} {Dataset} {Quality} {Enhancement}},
\newblock in: \bibinfo{editor}{B.~J. Choi}, \bibinfo{editor}{D.~Singh}, \bibinfo{editor}{U.~S. Tiwary}, \bibinfo{editor}{W.-Y. Chung} (Eds.), \bibinfo{booktitle}{Intelligent {Human} {Computer} {Interaction}}, \bibinfo{year}{2024}, pp. \bibinfo{pages}{295--309}. \DOIprefix\doi{10.1007/978-3-031-53827-8_27}.
\bibitem[{Stock and Cisse(2018)}]{stock_convnets_2018}
\bibinfo{author}{P.~Stock}, \bibinfo{author}{M.~Cisse},
\newblock \bibinfo{title}{{ConvNets} and {ImageNet} {Beyond} {Accuracy}: {Understanding} {Mistakes} and {Uncovering} {Biases}},
\newblock in: \bibinfo{booktitle}{The {European} {Conference} on {Computer} {Vision}}, \bibinfo{year}{2018}, pp. \bibinfo{pages}{504--519}. \DOIprefix\doi{10.1007/978-3-030-01231-1_31}.
\bibitem[{Engstrom et~al.(2020)Engstrom, Ilyas, Santurkar, Tsipras, Steinhardt, and Madry}]{engstrom_identifying_2020}
\bibinfo{author}{L.~Engstrom}, \bibinfo{author}{A.~Ilyas}, \bibinfo{author}{S.~Santurkar}, \bibinfo{author}{D.~Tsipras}, \bibinfo{author}{J.~Steinhardt}, \bibinfo{author}{A.~Madry},
\newblock \bibinfo{title}{Identifying {Statistical} {Bias} in {Dataset} {Replication}},
\newblock in: \bibinfo{booktitle}{Proceedings of the 37th {International} {Conference} on {Machine} {Learning}}, volume \bibinfo{volume}{119}, \bibinfo{year}{2020}, pp. \bibinfo{pages}{2922--2932}.
\bibitem[{Anzaku et~al.(2025)Anzaku, Wang, Babalola, Van~Messem, and De~Neve}]{anzaku_re-assessing_2025}
\bibinfo{author}{E.~T. Anzaku}, \bibinfo{author}{H.~Wang}, \bibinfo{author}{A.~Babalola}, \bibinfo{author}{A.~Van~Messem}, \bibinfo{author}{W.~De~Neve},
\newblock \bibinfo{title}{Re-assessing accuracy degradation: a framework for understanding {DNN} behavior on similar-but-non-identical test datasets},
\newblock \bibinfo{journal}{Machine Learning} \bibinfo{volume}{114} (\bibinfo{year}{2025}). \URLprefix \url{https://doi.org/10.1007/s10994-024-06693-x}. \DOIprefix\doi{10.1007/s10994-024-06693-x}.
\bibitem[{Taesiri et~al.(2023)Taesiri, Nguyen, Habchi, Bezemer, and Nguyen}]{taesiri_imagenet-hard_2023}
\bibinfo{author}{M.~R. Taesiri}, \bibinfo{author}{G.~Nguyen}, \bibinfo{author}{S.~Habchi}, \bibinfo{author}{C.-P. Bezemer}, \bibinfo{author}{A.~Nguyen},
\newblock \bibinfo{title}{{ImageNet}-{Hard}: {The} {Hardest} {Images} {Remaining} from a {Study} of the {Power} of {Zoom} and {Spatial} {Biases} in {Image} {Classification}},
\newblock in: \bibinfo{booktitle}{Conference on {Neural} {Information} {Processing} {Systems}}, \bibinfo{year}{2023}.
\bibitem[{Cole et~al.(2021)Cole, Aodha, Lorieul, Perona, Morris, and Jojic}]{cole_multi-label_2021}
\bibinfo{author}{E.~Cole}, \bibinfo{author}{O.~M. Aodha}, \bibinfo{author}{T.~Lorieul}, \bibinfo{author}{P.~Perona}, \bibinfo{author}{D.~Morris}, \bibinfo{author}{N.~Jojic},
\newblock \bibinfo{title}{Multi-{Label} {Learning} from {Single} {Positive} {Labels}},
\newblock in: \bibinfo{booktitle}{{IEEE}/{CVF} {Conference} on {Computer} {Vision} and {Pattern} {Recognition}}, \bibinfo{year}{2021}, pp. \bibinfo{pages}{933--942}. \DOIprefix\doi{10.1109/CVPR46437.2021.00099}.
\bibitem[{Verelst et~al.(2023)Verelst, Rubenstein, Eichner, Tuytelaars, and Berman}]{verelst_spatial_2023}
\bibinfo{author}{T.~Verelst}, \bibinfo{author}{P.~K. Rubenstein}, \bibinfo{author}{M.~Eichner}, \bibinfo{author}{T.~Tuytelaars}, \bibinfo{author}{M.~Berman},
\newblock \bibinfo{title}{Spatial {Consistency} {Loss} for {Training} {Multi}-{Label} {Classifiers} from {Single}-{Label} {Annotations}},
\newblock in: \bibinfo{booktitle}{{IEEE}/{CVF} {Winter} {Conference} on {Applications} of {Computer} {Vision}}, \bibinfo{year}{2023}, pp. \bibinfo{pages}{3868--3878}. \DOIprefix\doi{10.1109/WACV56688.2023.00387}.
\bibitem[{Yun et~al.(2021)Yun, Oh, Heo, Han, Choe, and Chun}]{yun_re-labeling_2021}
\bibinfo{author}{S.~Yun}, \bibinfo{author}{S.~J. Oh}, \bibinfo{author}{B.~Heo}, \bibinfo{author}{D.~Han}, \bibinfo{author}{J.~Choe}, \bibinfo{author}{S.~Chun},
\newblock \bibinfo{title}{Re-labeling {ImageNet}: from {Single} to {Multi}-{Labels}, from {Global} to {Localized} {Labels}},
\newblock in: \bibinfo{booktitle}{The {IEEE}/{CVF} {Conference} on {Computer} {Vision} and {Pattern} {Recognition}}, \bibinfo{year}{2021}, pp. \bibinfo{pages}{2340--2350}. \DOIprefix\doi{10.1109/CVPR46437.2021.00237}.
\bibitem[{Yun et~al.(2019)Yun, Han, Oh, Chun, Choe, and Yoo}]{yun_cutmix_2019}
\bibinfo{author}{S.~Yun}, \bibinfo{author}{D.~Han}, \bibinfo{author}{S.~J. Oh}, \bibinfo{author}{S.~Chun}, \bibinfo{author}{J.~Choe}, \bibinfo{author}{Y.~Yoo},
\newblock \bibinfo{title}{{CutMix}: {Regularization} {Strategy} to {Train} {Strong} {Classifiers} with {Localizable} {Features}},
\newblock in: \bibinfo{booktitle}{International {Conference} on {Computer} {Vision} ({ICCV})}, \bibinfo{year}{2019}.
\bibitem[{Takahashi et~al.(2018)Takahashi, Matsubara, and Uehara}]{takahashi_ricap_2018}
\bibinfo{author}{R.~Takahashi}, \bibinfo{author}{T.~Matsubara}, \bibinfo{author}{K.~Uehara},
\newblock \bibinfo{title}{{RICAP}: {Random} {Image} {Cropping} and {Patching} {Data} {Augmentation} for {Deep} {CNNs}},
\newblock in: \bibinfo{booktitle}{Proceedings of {The} 10th {Asian} {Conference} on {Machine} {Learning}}, volume~\bibinfo{volume}{95}, \bibinfo{year}{2018}, pp. \bibinfo{pages}{786--798}.
\bibitem[{Deng et~al.(2020)Deng, Liu, Shi, Zhang, Yang, and Liu}]{deng_deep_2020}
\bibinfo{author}{L.~Deng}, \bibinfo{author}{Y.~Liu}, \bibinfo{author}{Y.~Shi}, \bibinfo{author}{W.~Zhang}, \bibinfo{author}{C.~Yang}, \bibinfo{author}{H.~Liu},
\newblock \bibinfo{title}{Deep neural networks for inferring binding sites of {RNA}-binding proteins by using distributed representations of {RNA} primary sequence and secondary structure},
\newblock \bibinfo{journal}{BMC Genomics} \bibinfo{volume}{21} (\bibinfo{year}{2020}) \bibinfo{pages}{866}. \URLprefix \url{https://doi.org/10.1186/s12864-020-07239-w}. \DOIprefix\doi{10.1186/s12864-020-07239-w}.
\bibitem[{Zhang and Zhou(2014)}]{zhang_review_2014}
\bibinfo{author}{M.-L. Zhang}, \bibinfo{author}{Z.-H. Zhou},
\newblock \bibinfo{title}{A {Review} on {Multi}-{Label} {Learning} {Algorithms}},
\newblock \bibinfo{journal}{IEEE Transactions on Knowledge and Data Engineering} \bibinfo{volume}{26} (\bibinfo{year}{2014}) \bibinfo{pages}{1819--1837}. \DOIprefix\doi{10.1109/TKDE.2013.39}.
\bibitem[{Wightman(2019)}]{wightman_pytorch_2019}
\bibinfo{author}{R.~Wightman}, \bibinfo{title}{{PyTorch} {Image} {Models}}, \bibinfo{year}{2019}. \URLprefix \url{https://github.com/rwightman/pytorch-image-models}. \DOIprefix\doi{10.5281/zenodo.4414861}, \bibinfo{note}{publication Title: GitHub repository}.
\bibitem[{He et~al.(2016)He, Zhang, Ren, and Sun}]{he_deep_2016}
\bibinfo{author}{K.~He}, \bibinfo{author}{X.~Zhang}, \bibinfo{author}{S.~Ren}, \bibinfo{author}{J.~Sun},
\newblock \bibinfo{title}{Deep {Residual} {Learning} for {Image} {Recognition}},
\newblock in: \bibinfo{booktitle}{{IEEE} {Conference} on {Computer} {Vision} and {Pattern} {Recognition}}, \bibinfo{year}{2016}, pp. \bibinfo{pages}{770--778}. \DOIprefix\doi{10.1109/CVPR.2016.90}.
\bibitem[{Howard et~al.(2017)Howard, Zhu, Chen, Kalenichenko, Wang, Weyand, Andreetto, and Adam}]{howard_mobilenets_2017}
\bibinfo{author}{A.~G. Howard}, \bibinfo{author}{M.~Zhu}, \bibinfo{author}{B.~Chen}, \bibinfo{author}{D.~Kalenichenko}, \bibinfo{author}{W.~Wang}, \bibinfo{author}{T.~Weyand}, \bibinfo{author}{M.~Andreetto}, \bibinfo{author}{H.~Adam}, \bibinfo{title}{{MobileNets}: {Efficient} {Convolutional} {Neural} {Networks} for {Mobile} {Vision} {Applications}}, \bibinfo{year}{2017}. \URLprefix \url{http://arxiv.org/abs/1704.04861}. \DOIprefix\doi{10.48550/arXiv.1704.04861}.
\bibitem[{Tan and Le(2019)}]{tan_efficientnet_2019}
\bibinfo{author}{M.~Tan}, \bibinfo{author}{Q.~V. Le},
\newblock \bibinfo{title}{{EfficientNet}: {Rethinking} {Model} {Scaling} for {Convolutional} {Neural} {Networks}},
\newblock in: \bibinfo{booktitle}{Proceedings of the 36th {International} {Conference} on {Machine} {Learning}}, volume~\bibinfo{volume}{97}, \bibinfo{year}{2019}, pp. \bibinfo{pages}{6105--6114}.
\bibitem[{Woo et~al.(2023)Woo, Debnath, Hu, Chen, Liu, Kweon, and Xie}]{woo_convnext_2023}
\bibinfo{author}{S.~Woo}, \bibinfo{author}{S.~Debnath}, \bibinfo{author}{R.~Hu}, \bibinfo{author}{X.~Chen}, \bibinfo{author}{Z.~Liu}, \bibinfo{author}{I.~S. Kweon}, \bibinfo{author}{S.~Xie},
\newblock \bibinfo{title}{{ConvNeXt} {V2}: {Co}-designing and {Scaling} {ConvNets} with {Masked} {Autoencoders}},
\newblock in: \bibinfo{booktitle}{{IEEE}/{CVF} {Conference} on {Computer} {Vision} and {Pattern} {Recognition}}, \bibinfo{year}{2023}, pp. \bibinfo{pages}{16133--16142}. \DOIprefix\doi{10.1109/CVPR52729.2023.01548}.
\bibitem[{Dosovitskiy et~al.(2021)Dosovitskiy, Beyer, Kolesnikov, Weissenborn, Zhai, Unterthiner, Dehghani, Minderer, Heigold, Gelly, Uszkoreit, and Houlsby}]{dosovitskiy_image_2021}
\bibinfo{author}{A.~Dosovitskiy}, \bibinfo{author}{L.~Beyer}, \bibinfo{author}{A.~Kolesnikov}, \bibinfo{author}{D.~Weissenborn}, \bibinfo{author}{X.~Zhai}, \bibinfo{author}{T.~Unterthiner}, \bibinfo{author}{M.~Dehghani}, \bibinfo{author}{M.~Minderer}, \bibinfo{author}{G.~Heigold}, \bibinfo{author}{S.~Gelly}, \bibinfo{author}{J.~Uszkoreit}, \bibinfo{author}{N.~Houlsby},
\newblock \bibinfo{title}{An {Image} is {Worth} 16x16 {Words}: {Transformers} for {Image} {Recognition} at {Scale}},
\newblock in: \bibinfo{booktitle}{Ninth {International} {Conference} on {Learning} {Representations}}, \bibinfo{year}{2021}.
\bibitem[{Touvron et~al.(2021)Touvron, Cord, Douze, Massa, Sablayrolles, and Jégou}]{touvron_training_2021}
\bibinfo{author}{H.~Touvron}, \bibinfo{author}{M.~Cord}, \bibinfo{author}{M.~Douze}, \bibinfo{author}{F.~Massa}, \bibinfo{author}{A.~Sablayrolles}, \bibinfo{author}{H.~Jégou},
\newblock \bibinfo{title}{Training {Data}-efficient {Image} {Transformers} \& {Distillation} {Through} {Attention}},
\newblock in: \bibinfo{booktitle}{Proceedings of the 38th {International} {Conference} on {Machine} {Learning}}, volume \bibinfo{volume}{139}, \bibinfo{year}{2021}, pp. \bibinfo{pages}{10347--10357}.
\bibitem[{Dong et~al.(2022)Dong, Piao, and Wei}]{dong_beit_2022}
\bibinfo{author}{L.~Dong}, \bibinfo{author}{S.~Piao}, \bibinfo{author}{F.~Wei},
\newblock \bibinfo{title}{{BEiT}: {BERT} {Pre}-{Training} of {Image} {Transformers}},
\newblock in: \bibinfo{booktitle}{The {Tenth} {International} {Conference} on {Learning} {Representations}}, \bibinfo{year}{2022}.
\bibitem[{Liu et~al.(2021)Liu, Lin, Cao, Hu, Wei, Zhang, Lin, and Guo}]{liu_swin_2021}
\bibinfo{author}{Z.~Liu}, \bibinfo{author}{Y.~Lin}, \bibinfo{author}{Y.~Cao}, \bibinfo{author}{H.~Hu}, \bibinfo{author}{Y.~Wei}, \bibinfo{author}{Z.~Zhang}, \bibinfo{author}{S.~Lin}, \bibinfo{author}{B.~Guo},
\newblock \bibinfo{title}{Swin {Transformer}: {Hierarchical} {Vision} {Transformer} using {Shifted} {Windows}},
\newblock in: \bibinfo{booktitle}{{IEEE}/{CVF} {International} {Conference} on {Computer} {Vision}}, \bibinfo{year}{2021}, pp. \bibinfo{pages}{9992--10002}. \DOIprefix\doi{10.1109/ICCV48922.2021.00986}.
\bibitem[{Touvron et~al.(2021)Touvron, Cord, Sablayrolles, Synnaeve, and Jegou}]{touvron_going_2021}
\bibinfo{author}{H.~Touvron}, \bibinfo{author}{M.~Cord}, \bibinfo{author}{A.~Sablayrolles}, \bibinfo{author}{G.~Synnaeve}, \bibinfo{author}{H.~Jegou},
\newblock \bibinfo{title}{Going deeper with {Image} {Transformers}},
\newblock in: \bibinfo{booktitle}{{IEEE}/{CVF} {International} {Conference} on {Computer} {Vision} ({ICCV})}, \bibinfo{publisher}{IEEE}, \bibinfo{year}{2021}, pp. \bibinfo{pages}{32--42}. \DOIprefix\doi{10.1109/ICCV48922.2021.00010}.
\bibitem[{Yuan et~al.(2023)Yuan, Hou, Jiang, Feng, and Yan}]{yuan_volo_2023}
\bibinfo{author}{L.~Yuan}, \bibinfo{author}{Q.~Hou}, \bibinfo{author}{Z.~Jiang}, \bibinfo{author}{J.~Feng}, \bibinfo{author}{S.~Yan},
\newblock \bibinfo{title}{{VOLO}: {Vision} {Outlooker} for {Visual} {Recognition}},
\newblock \bibinfo{journal}{IEEE Transactions on Pattern Analysis and Machine Intelligence} \bibinfo{volume}{45} (\bibinfo{year}{2023}) \bibinfo{pages}{6575--6586}. \URLprefix \url{https://ieeexplore.ieee.org/document/9888055}. \DOIprefix\doi{10.1109/TPAMI.2022.3206108}.
\bibitem[{Sun et~al.(2023)Sun, Fang, Wu, Wang, and Cao}]{sun_eva-clip_2023}
\bibinfo{author}{Q.~Sun}, \bibinfo{author}{Y.~Fang}, \bibinfo{author}{L.~Wu}, \bibinfo{author}{X.~Wang}, \bibinfo{author}{Y.~Cao}, \bibinfo{title}{{EVA}-{CLIP}: {Improved} {Training} {Techniques} for {CLIP} at {Scale}}, \bibinfo{year}{2023}. \DOIprefix\doi{10.48550/arXiv.2303.15389}.
\bibitem[{Mahajan et~al.(2018)Mahajan, He, Paluri, Li, Bharambe, and Van Der~Maaten}]{mahajan_exploring_2018}
\bibinfo{author}{D.~Mahajan}, \bibinfo{author}{K.~He}, \bibinfo{author}{M.~Paluri}, \bibinfo{author}{Y.~Li}, \bibinfo{author}{A.~Bharambe}, \bibinfo{author}{L.~Van Der~Maaten},
\newblock \bibinfo{title}{Exploring the {Limits} of {Weakly} {Supervised} {Pretraining}},
\newblock in: \bibinfo{booktitle}{European {Conference} on {Computer} {Vision}}, volume \bibinfo{volume}{11206}, \bibinfo{year}{2018}, pp. \bibinfo{pages}{185--201}.
\bibitem[{Singh et~al.(2022)Singh, Gustafson, Adcock, De~Freitas~Reis, Gedik, Kosaraju, Mahajan, Girshick, Dollar, and Van Der~Maaten}]{singh_revisiting_2022}
\bibinfo{author}{M.~Singh}, \bibinfo{author}{L.~Gustafson}, \bibinfo{author}{A.~Adcock}, \bibinfo{author}{V.~De~Freitas~Reis}, \bibinfo{author}{B.~Gedik}, \bibinfo{author}{R.~P. Kosaraju}, \bibinfo{author}{D.~Mahajan}, \bibinfo{author}{R.~Girshick}, \bibinfo{author}{P.~Dollar}, \bibinfo{author}{L.~Van Der~Maaten},
\newblock \bibinfo{title}{Revisiting {Weakly} {Supervised} {Pre}-{Training} of {Visual} {Perception} {Models}},
\newblock in: \bibinfo{booktitle}{{IEEE}/{CVF} {Conference} on {Computer} {Vision} and {Pattern} {Recognition}}, \bibinfo{year}{2022}, pp. \bibinfo{pages}{794--804}. \DOIprefix\doi{10.1109/CVPR52688.2022.00088}.
\bibitem[{Jiang et~al.(2021)Jiang, Hou, Yuan, Zhou, Shi, Jin, Wang, and Feng}]{jiang_all_2021}
\bibinfo{author}{Z.-H. Jiang}, \bibinfo{author}{Q.~Hou}, \bibinfo{author}{L.~Yuan}, \bibinfo{author}{D.~Zhou}, \bibinfo{author}{Y.~Shi}, \bibinfo{author}{X.~Jin}, \bibinfo{author}{A.~Wang}, \bibinfo{author}{J.~Feng},
\newblock \bibinfo{title}{All {Tokens} {Matter}: {Token} {Labeling} for {Training} {Better} {Vision} {Transformers}},
\newblock in: \bibinfo{booktitle}{Advances in {Neural} {Information} {Processing} {Systems}}, volume~\bibinfo{volume}{34}, \bibinfo{year}{2021}, pp. \bibinfo{pages}{18590--18602}.
\bibitem[{Liu et~al.(2022)Liu, Mao, Wu, Feichtenhofer, Darrell, and Xie}]{liu_convnet_2022}
\bibinfo{author}{Z.~Liu}, \bibinfo{author}{H.~Mao}, \bibinfo{author}{C.-Y. Wu}, \bibinfo{author}{C.~Feichtenhofer}, \bibinfo{author}{T.~Darrell}, \bibinfo{author}{S.~Xie},
\newblock \bibinfo{title}{A {ConvNet} for the 2020s},
\newblock in: \bibinfo{booktitle}{{IEEE} {Conference} on {Computer} {Vision} and {Pattern} {Recognition}}, \bibinfo{year}{2022}, pp. \bibinfo{pages}{11976--11986}.
\bibitem[{Peng et~al.(2022)Peng, Dong, Bao, Ye, and Wei}]{peng_beit_2022}
\bibinfo{author}{Z.~Peng}, \bibinfo{author}{L.~Dong}, \bibinfo{author}{H.~Bao}, \bibinfo{author}{Q.~Ye}, \bibinfo{author}{F.~Wei}, \bibinfo{title}{{BEiT} v2: {Masked} {Image} {Modeling} with {Vector}-{Quantized} {Visual} {Tokenizers}}, \bibinfo{year}{2022}. \URLprefix \url{http://arxiv.org/abs/2208.06366}. \DOIprefix\doi{10.48550/arXiv.2208.06366}.
\bibitem[{Radford et~al.(2021)Radford, Kim, Hallacy, Ramesh, Goh, Agarwal, Sastry, Askell, Mishkin, Clark, Krueger, and Sutskever}]{radford_learning_2021}
\bibinfo{author}{A.~Radford}, \bibinfo{author}{J.~W. Kim}, \bibinfo{author}{C.~Hallacy}, \bibinfo{author}{A.~Ramesh}, \bibinfo{author}{G.~Goh}, \bibinfo{author}{S.~Agarwal}, \bibinfo{author}{G.~Sastry}, \bibinfo{author}{A.~Askell}, \bibinfo{author}{P.~Mishkin}, \bibinfo{author}{J.~Clark}, \bibinfo{author}{G.~Krueger}, \bibinfo{author}{I.~Sutskever},
\newblock \bibinfo{title}{Learning {Transferable} {Visual} {Models} {From} {Natural} {Language} {Supervision}},
\newblock in: \bibinfo{booktitle}{Proceedings of the 38th {International} {Conference} on {Machine} {Learning}}, \bibinfo{year}{2021}, pp. \bibinfo{pages}{8748--8763}.
\bibitem[{Radosavovic et~al.(2020)Radosavovic, Kosaraju, Girshick, He, and Dollar}]{radosavovic_designing_2020}
\bibinfo{author}{I.~Radosavovic}, \bibinfo{author}{R.~P. Kosaraju}, \bibinfo{author}{R.~Girshick}, \bibinfo{author}{K.~He}, \bibinfo{author}{P.~Dollar},
\newblock \bibinfo{title}{Designing {Network} {Design} {Spaces}},
\newblock in: \bibinfo{booktitle}{{IEEE}/{CVF} {Conference} on {Computer} {Vision} and {Pattern} {Recognition} ({CVPR})}, \bibinfo{year}{2020}, pp. \bibinfo{pages}{10425--10433}. \DOIprefix\doi{10.1109/CVPR42600.2020.01044}.
\bibitem[{Kruthiventi et~al.(2015)Kruthiventi, Ayush, and Babu}]{kruthiventi_deepfix_2015}
\bibinfo{author}{S.~S.~S. Kruthiventi}, \bibinfo{author}{K.~Ayush}, \bibinfo{author}{R.~V. Babu}, \bibinfo{title}{{DeepFix}: {A} {Fully} {Convolutional} {Neural} {Network} for predicting {Human} {Eye} {Fixations}}, \bibinfo{year}{2015}. \URLprefix \url{http://arxiv.org/abs/1510.02927}. \DOIprefix\doi{10.48550/arXiv.1510.02927}.
\bibitem[{Jetley et~al.(2016)Jetley, Murray, and Vig}]{jetley_end--end_2016}
\bibinfo{author}{S.~Jetley}, \bibinfo{author}{N.~Murray}, \bibinfo{author}{E.~Vig},
\newblock \bibinfo{title}{End-to-{End} {Saliency} {Mapping} via {Probability} {Distribution} {Prediction}},
\newblock in: \bibinfo{booktitle}{{IEEE} {Conference} on {Computer} {Vision} and {Pattern} {Recognition}}, \bibinfo{year}{2016}, pp. \bibinfo{pages}{5753--5761}. \DOIprefix\doi{10.1109/CVPR.2016.620}.

\end{thebibliography}
\end{document}